%% file: arxiv.tex
\documentclass[10pt, letterpaper, twoside]{article}

\usepackage[margin=1.5in]{geometry}
\usepackage[round]{natbib}
\usepackage{pratik}

\input{./notation}

\graphicspath{{../fig/},{../}}

\title{Deep Reference Priors:\\What is the best way to pretrain a model?
\footnote{Proceedings of the 39$^{\text{th}}$ International Conference on Machine Learning, Baltimore, Maryland, USA. Copyright 2022 by the authors.}}
\author[1,\textdagger]{Yansong Gao}
\author[2,\textdagger]{Rahul Ramesh}
\author[2,3]{Pratik Chaudhari}

\affil[1]{\normalsize Applied Mathematics and Computational Sciences, University of Pennsylvania\vspace*{0.25em}}
\affil[2]{\normalsize Computer and Information Science, University of Pennsylvania\vspace*{0.25em}}
\affil[3]{\normalsize Electrical and Systems Engineering, University of Pennsylvania\vspace*{0.25em}}
\affil[]{\normalsize \textdagger Equal contribution}
\affil[ ]{\normalsize Email: \href{mailto:gaoyans@sas.upenn.edu}{gaoyans@sas.upenn.edu}, \href{mailto:rahulram@seas.upenn.edu}{rahulram@seas.upenn.edu}, \href{mailto:pratikac@seas.upenn.edu}{pratikac@seas.upenn.edu}}
\date{}

\begin{document}
\maketitle

\input{abstract}
\input{intro}
\input{background}
\input{methods}

\input{experiments}

\input{related_work}

\section{Acknowledgments}
This work was supported by grants from the National Science Foundation (2145164) and the Office of Naval Research (N00014-22-1-2255), and cloud computing credits from Amazon Web Services.

\begin{small}
\setlength{\bibsep}{1em}
\bibliography{main}
\end{small}

\input{appendix}

\end{document}

%% file: notation.tex
\def \mid {\,|\,}

\def \ps {P^s}
\def \pt {P^t}
\def \hpn {\hat{P}_n}

\def \xm {x^m}
\def \ym {y^m}
\def \zm {z^m}
\def \xn {x^n}
\def \yn {y^n}
\def \zn {z^n}

\def \xu {x^u}
\def \yu {y^u}

\def \zmn {z^{m+n}}

\def \xms {\xm_s}
\def \yms {\ym_s}
\def \xnt {\xn_t}
\def \ynt {\yn_t}

\def \KL {\text{KL}}

\def \pin {\pi_n}
\def \pinm {\pi_{n \mid m}}

%% file: abstract.tex
% !TEX root = ./ICML_main.tex

\begin{abstract}
What is the best way to exploit extra data---be it unlabeled data from the same task, or labeled data from a related task---to learn a given task? This paper formalizes the question using the theory of reference priors. Reference priors are objective, uninformative Bayesian priors that maximize the mutual information between the task and the weights of the model. Such priors enable the task to maximally affect the Bayesian posterior, e.g., reference priors depend upon the number of samples available for learning the task and for very small sample sizes, the prior puts more probability mass on low-complexity models in the hypothesis space. This paper presents the first demonstration of reference priors for medium-scale deep networks and image-based data. We develop generalizations of reference priors and demonstrate applications to two problems. First, by using unlabeled data to compute the reference prior, we develop new Bayesian semi-supervised learning methods that remain effective even with very few samples per class. Second, by using labeled data from the source task to compute the reference prior, we develop a new pretraining method for transfer learning that allows data from the target task to maximally affect the Bayesian posterior. Empirical validation of these methods is conducted on image classification datasets. Code is available at \href{https://github.com/grasp-lyrl/deep_reference_priors}{https://github.com/grasp-lyrl/deep\_reference\_priors}.
\end{abstract}

%% file: intro.tex
% !TEX root = ./ICML_main.tex

\section{Introduction}
\label{s:intro}

Exploiting extra data, e.g., labeled data from a related task, or unlabeled data from the same task, is a powerful way of reducing the number of training data required to learn a given task. This idea lies at the heart of burgeoning fields like transfer, meta-, semi- and self-supervised learning, and these fields have developed a wide variety of methods to incorporate such extra information. To give a few examples, methods for transfer learning fine-tune a representation that was pretrained on labeled data from another---ideally related---task. Methods for semi-supervised learning pretrain the representation using unlabeled data, which may come from the same task or from other related tasks, before using the labeled data. In this paper, we ask the question: what is the \emph{best} way to exploit extra data for learning a task? In other words, if we have \emph{some} pool of data---be it labeled or unlabeled, from the same task, or from another task---what is the \emph{optimal} way to pretrain a representation?

As posed, the answer to the question above depends upon the downstream task that we seek to solve. But we can ask a more reasonable question by recognizing that a pretrained representation can be thought of as a Bayesian prior (or a sample from it). Fundamentally, a prior restricts the set of models that can be fitted upon the task. So we could instead ask: \emph{how to best use the extra data to restrict the set of models that we could fit on the desired task}. This paper formalizes the question using the concept of reference priors and makes the following contributions.

\begin{enumerate}[(1), wide,labelwidth=0ex, labelindent=\parindent]
\item We \textbf{formalize the problem of ``how to best pretrain a model''} using the theory of reference priors, which are objective, uninformative Bayesian priors computed by maximizing the mutual information between the task and the weights. We show how these priors maximize the KL-divergence between the posterior computed from the task and the prior, on average over the distribution of the unknown future data. This allows the samples from the task to maximally influence the posterior. We discuss how reference priors are supported on a discrete set of atoms in the weight space. We \textbf{develop a method to compute reference priors for deep networks}. To our knowledge, this is the \textbf{first instantiation of reference priors for deep networks that preserves their characteristic discrete nature}.

\item We \textbf{formalize semi-supervised learning as computing a reference prior} where the learner is given access to a pool of unlabeled data and seeks to compute a prior using this data. This formulation sheds light upon the \textbf{theoretical underpinnings of existing state of the art methods such as FixMatch}. We show that techniques such as consistency regularization and entropy minimization which are commonly used in practice can be directly understood using the reference prior formulation.

\item We \textbf{formalize transfer learning as building a two-stage reference prior} where the learner gets access to data in two stages and computes a prior that is optimal for data from the second stage. Such a prior has the flavor of ignoring certain parts of the weight space depending upon whether data from the first stage was similar to that from the second stage, or not. This formulation is useful because it is an information-theoretically optimal way to pretrain using a source task for the goal of transferring to the target task. This objective is closely related to the predictive Information Bottleneck principle.

\item We show an empirical study of our formulations on the CIFAR-10 and CIFAR-100 datasets. We show that \textbf{our methods to compute reference priors provide results that are competitive with state of the art} methods for semi-supervised learning, e.g., we obtain an \textbf{accuracy of 85.45\% on CIFAR-10 with 5 labeled samples/class}. We obtain significantly better accuracy than well-tuned fine-tuning for transfer learning, even for very small sample sizes.
\end{enumerate}

%% file: background.tex
% !TEX root = ./ICML_main.tex

\section{Background}
\label{s:background}

%This section introduces notation, develops the formulation for objective Bayesian priors and gives a few examples that explain how these priors work.

\subsection{Setup}

Consider a dataset $\hpn = \cbr{(x_i, y_i)}_{i=1}^n$ with $n$ samples that consists of inputs $x_i \in \reals^d$ and labels $y_i \in \cbr{1,\ldots,C}$. Each sample of this dataset is drawn from a joint distribution $P(x, y)$ which we define to be the ``task''.  We will use the shorthand $\xn = (x_1,\ldots,x_n)$ and $\yn = (y_1,\ldots,y_n)$ to denote all inputs and labels. Let $w \in \reals^p$ be the weights of a probabilistic model which evaluates $p_w(y  \mid x)$. We will use a random variable $z$ with a probabilistic model $p_w(z)$ when we do not wish to distinguish between inputs and labels.

Given a prior on weights $\pi(w)$, Bayes law gives the posterior
\(
    p(w \mid \xn, \yn) \propto p(\yn \mid \xn, w) \pi(w).
\)
The Fisher Information Matrix (FIM) $g \in \reals^{p \times p}$ has entries $g(w)_{kl}=$
\[
    \f 1 n \sum_{i=1}^n \sum_{y=1}^C p_w(y \mid x_i) \partial_{w_k} \log p_w(y \mid x_i) \partial_{w_l} \log p_w(y \mid x_i).
\]
It can be used to define the Jeffreys prior $\pi_J(w) \propto \sqrt{\det g(w)}$. Jeffreys prior is reparameterization invariant, i.e., it assigns the same probability to a set of models irrespective of our choice of parameterization of those models. It is an uninformative prior, e.g., it imposes some generic structure on the problem (reparameterization invariance).
%Informative priors are different from such uninformative priors. For instance, if we were to choose $\pi$ to be a Gaussian distribution with mean at a pretrained model, then such a prior would express a very definitive information and bias the posterior towards the pretrained model.
\subsection{Reference Priors}
\label{s:reference_priors}

To make the choice of a prior more objective, \citet{bernardo1979reference} suggested that uninformative priors should maximize some divergence, say the Kullback-Leibler (KL) divergence $\KL(p(w \mid z), \pi(w)) = \int \dd{w} p(w \mid z) \log \rbr{p(w \mid z)/\pi(w)}$, between the prior and the posterior for data $z$. The rationale for doing so is to allow the data to dominate the posterior rather than our choice of the prior. Since we do not know the data \emph{a priori} while picking the prior, we should maximize the \emph{average} KL-divergence over the data distribution $p(z)$. This amounts to maximizing the mutual information
\beq{
    \aed{
        &\pi^* = \argmax_\pi  I_\pi(w; z) \\
        &:= \int \dd{z} \dd{w} p(z) p(w \mid z) \log \f{p(w \mid z)}{\pi(w)}
        = H(w) - H(w \mid z)
        %&= \int \dd{z} \dd{w} p(z) p(w \mid z) \log \f{p(z \mid w)}{p(z)}
%        &= \int \dd{z} \dd{w} \pi(w) p(z \mid w) \log \f{p(z \mid w)}{p(z)}
 %       = H(z) - H(z \mid w),
    }
    \label{eq:ref_defn}
}
where $p(z) = \int \dd{w} \pi(w) p(z \mid w)$ and $H(w) = - \int \dd{w} \pi(w) \log \pi(w)$ is the Shannon entropy; the conditional entropy $H(w \mid z)$ is defined analogously. Mutual information is a natural quantity for measuring the amount of missing information about $w$ provided by data $z$ if the initial belief was $\pi$. The prior $\pi^*(w)$ is known as a reference prior. It is invariant to a reparameterization of the weight space because mutual information is invariant to reparameterization. The reference prior does not depend upon the samples $\hpn$ but only depends on their distribution $P$.

The objective to calculate reference prior $\pi^*$ above may not be analytically tractable and therefore Bernardo also suggested computing $n$-reference priors. We call $n$ the ``order'' and deliberately overload the notation for the number of samples $n$; the reason will be clear soon.
\beq{
    \aed{
        \textstyle \pin^* = \argmax_\pi  I_\pi(w; \zn)
        % &= \int \dd{\zn} \dd{w} p(\zn) p(w \mid \zn) \log \f{p(w \mid \zn)}{\pi(w)},\\
        &= H(w) - H(w \mid \zn),
    }
    \label{eq:ref_pin}
}
using $n$ samples and then setting $\pi^* := \lim_{n \to \infty} \pin^*$ under appropriate technical conditions~\citep{berger1988priors}. Reference priors are asymptotically equivalent to Jeffreys prior for one-dimensional problems. In general, they differ for multi-dimensional problems but it can be shown that Jeffreys prior is the continuous prior that maximizes the mutual information~\citep{clarke1994jeffreys}.

\subsection{Blahut-Arimoto algorithm}
\label{s:blahut_arimoto}

The Blahut-Arimoto algorithm~\citep{arimoto1972algorithm,blahut1972computation} is a method for maximizing functionals like~\cref{eq:ref_defn} and leads to iterations of the form $\pi^{t+1}(w) \propto \exp \rbr{\KL(p(z \mid w), p(z))} \pi^{t}(w)$.
% It exploits the identity
% \[
%     \max_\pi I_\pi(w; z) = \max_q \max_\pi \E_{w \sim \pi} \sbr{p(z \mid w) \log \f{q(w \mid z)}{\pi(w)}}
% \]
% where the optimum on the right-hand-side is achieved when $q(w \mid z) \equiv p(w \mid z)$. The BA algorithm alternates the maximization over $\pi$ and $q$ in this identity.
It is typically implemented for discrete variables, e.g., in the Information Bottleneck~\citep{tishbyInformationBottleneckMethod1999}. In this case, maximizing mutual information is a convex problem and therefore the BA algorithm is guaranteed to converge. Such discretization is difficult for high-dimensional deep networks. We therefore implement the BA algorithm using particles; see~\cref{rem:ba_particles}.

\begin{figure}[htpb]
\centering
\includegraphics[width=0.325\linewidth]{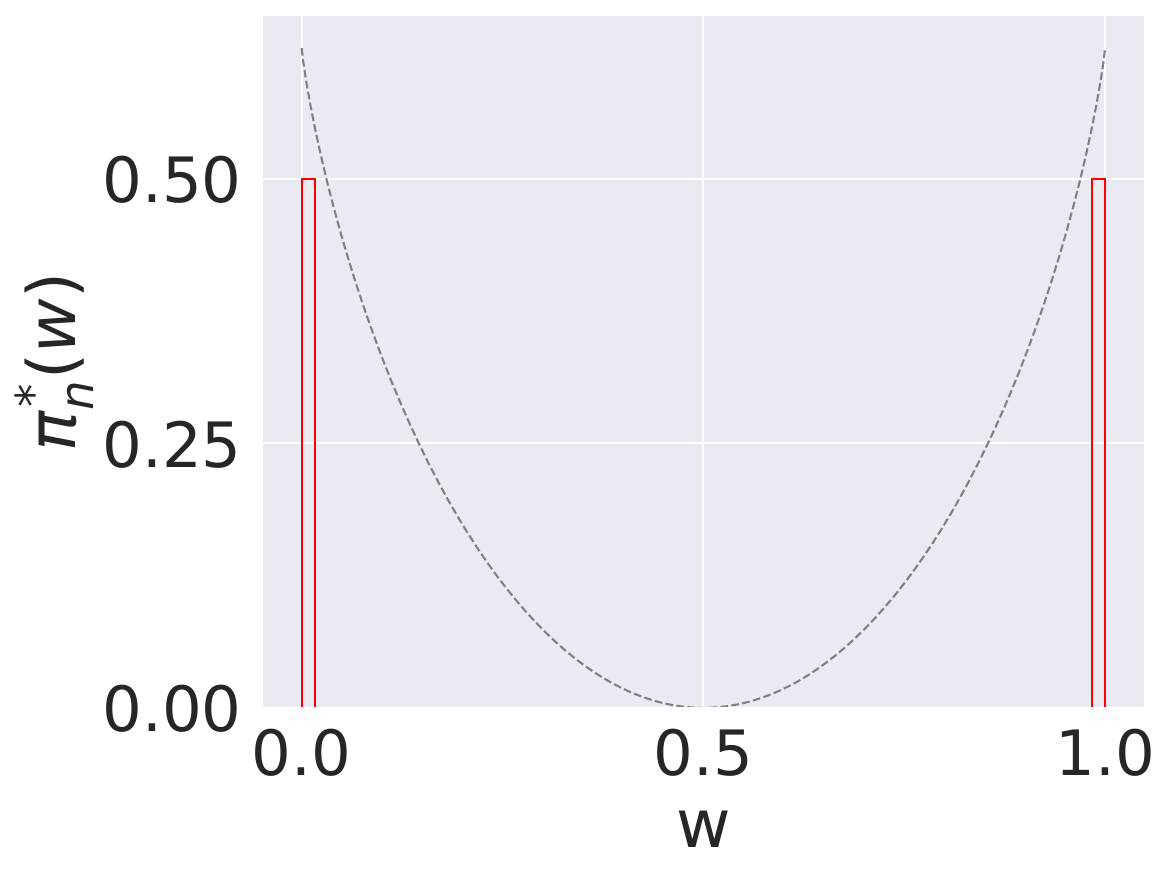}
\includegraphics[width=0.325\linewidth]{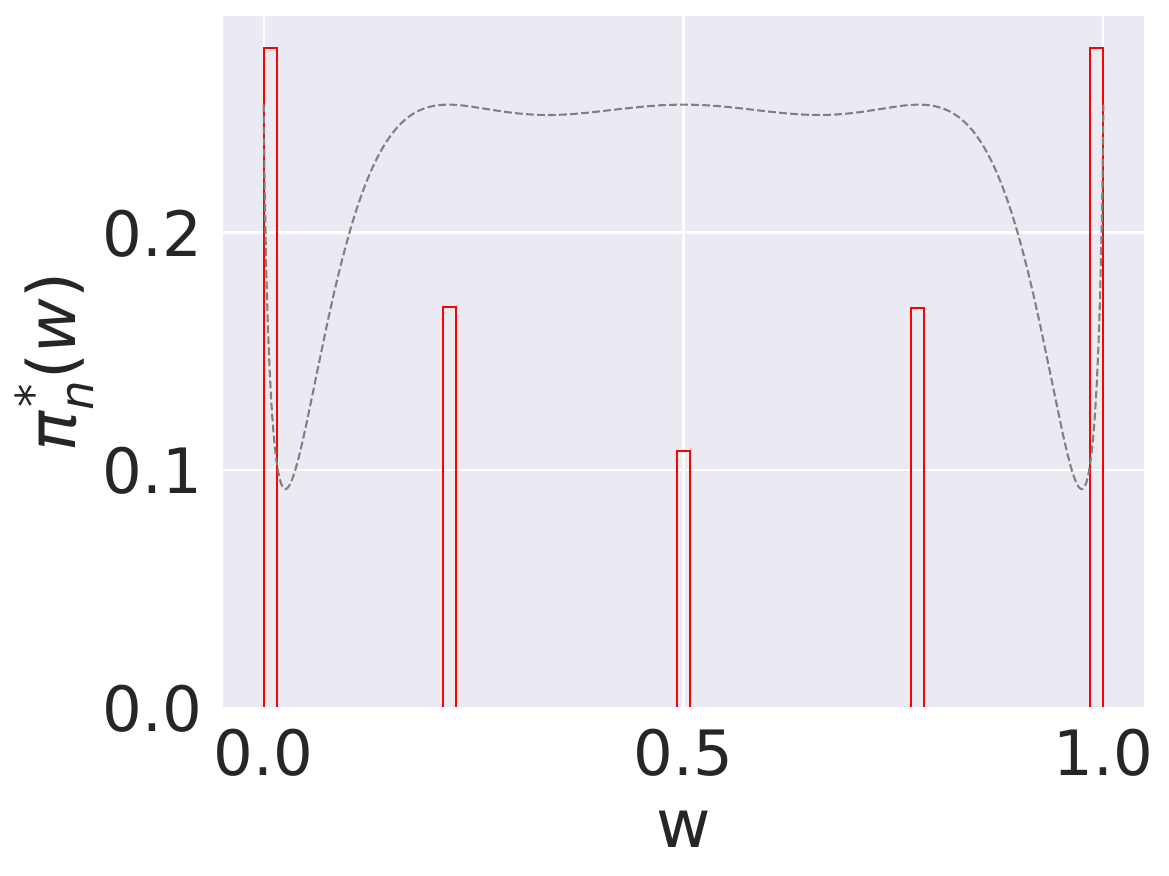}
\includegraphics[width=0.325\linewidth]{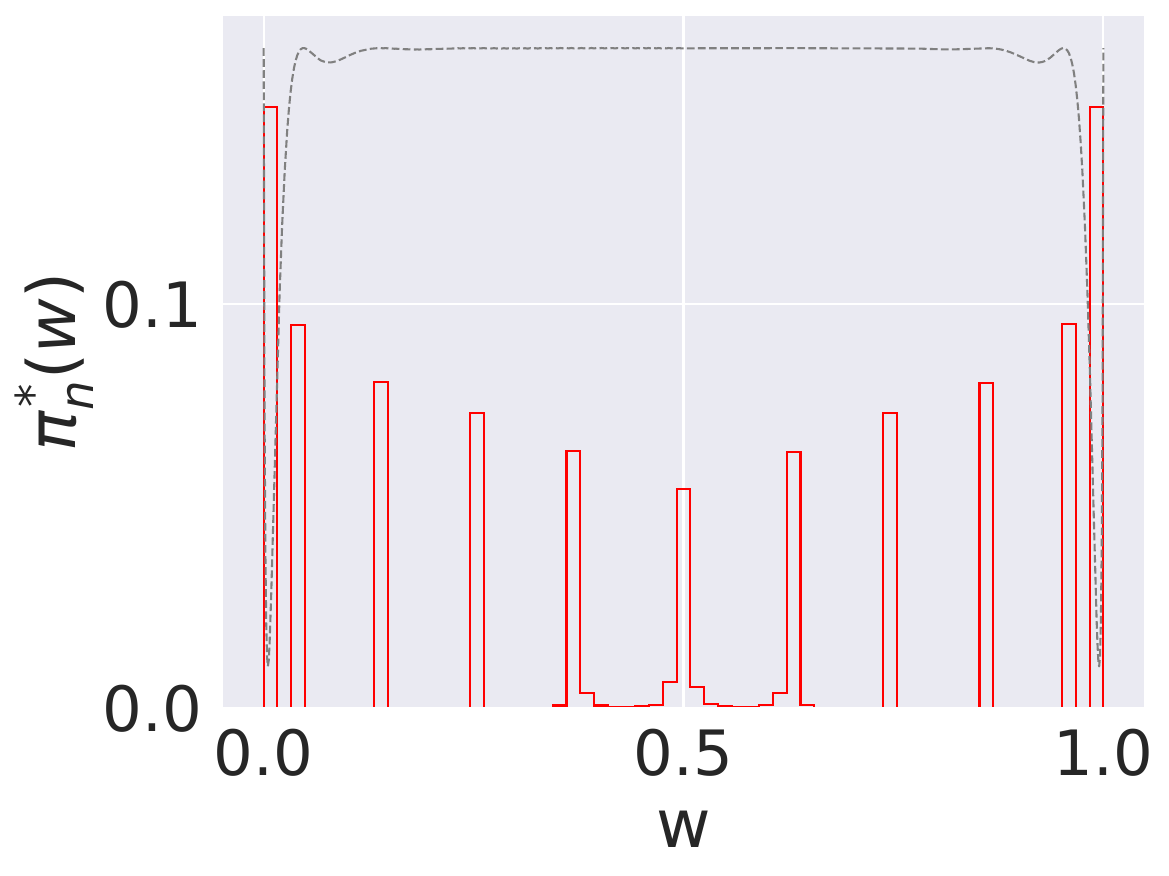}
\caption{We calculated the \textbf{reference prior for the coin-tossing model} for $n=1, 10, 50$ (from left to right) using the Blahut-Arimoto algorithm. Atoms are critical points of the gray line which is $\KL(p(\zn), p(\zn \mid w))$. The prior is discrete for finite order $n < \infty$~\citep{mattingly2018maximizing}. Atoms of the prior are maximally different from each other, e.g., for $n=1$, they are on opposite corners of the parameter space. As the number of samples increases, the separation between atoms of the prior reduces. The prior converges to Jeffreys prior $\pi_J(w) \propto \rbr{w (1-w)}^{-1}$ as $n \to \infty$.}
\label{fig:coin_convergence_jeffreys}
\end{figure}

\begin{example}[Estimating the bias of a coin]
\label{eg:bias}
To ground intuition, consider the estimation of the bias of a coin $w \in [0,1]$ using $n$ trials. If $z$ denotes the number of heads (which is a sufficient statistic),  we have $p(z \mid w) = w^z (1-w)^{n-z} n!/(z! (n-z)!)$.
%which gives Jeffreys prior $\pi_J(w) \propto \rbr{w (1-w)}^{-1}$.
For $n=1$, since we know that $I(w; z^1) \leq \log 2$ with this one bit of information, we can see that
\(
    \pi^*_1(z) = (\delta(w) + \delta(1-w))/2
\)
is the reference prior that achieves this upper bound. This result is intuitive: if we \emph{know} that we have only one observation, then the optimal uninformative prior should put equal probability mass on the two exhaustive outcomes $w=0$ (heads) and $w=1$ (tails). We can numerically calculate $\pin^*$ for different values of $n$ using the BA algorithm (\cref{fig:coin_convergence_jeffreys}).
\end{example}

%% file: methods.tex
% !TEX root = ./ICML_main.tex

\section{Methods}
\label{s:methods}

This section discusses a key property of reference priors that enables us to calculate them numerically, namely that they are supported on a discrete set in the weight space (\cref{s:discrete}). It then formulates reference priors for semi-supervised (\cref{s:ssl}) and transfer learning (\cref{s:two_stage,s:transfer}).

\subsection{Existence and discreteness of reference priors}
\label{s:existence}
\label{s:discrete}

Rigorous theoretical development of reference priors has been done in the statistics literature. We focus on their applications. We however mention some technical conditions under which our development remains meaningful.

A reference prior does not exist if $I_\pi(w; \zn)$ is infinite~\citep{berger1988priors}. For the concept of a reference prior to remain meaningful, we make the following technical assumptions. (i) $\pi$ is supported on a compact set $\Omega \subset \reals^p$, and (ii) if $p_{\pi}( \zn) = \int_{\Omega} \dd{w} \pi(w) p(\zn \mid w)$ is the marginal, then $\KL( p_w, p_\pi )$ is a continuous function of $w$ for any $\pi$. Under these conditions, the $n$-order prior $\pin^*$ exists and $I_{\pin}(w; \zn)$ is finite; see~\citep[Lemma 2.14]{zhang1994discrete}.
Now assume that $\pin^*$ exists and is unique up to a set of measure zero. Let $\Omega_n = \{ w \in \Omega: \pin^*(w) > 0 \}$ be the support of $\pin^*$ and $\zn$ be a discrete random variable with $C$ atoms. If $\{ p(\zn \mid w): w \in \Omega_n\}$ is compact, then $\pin^*$ is discrete with no more than $C$ atoms~\citep[Lemma 2.18]{zhang1994discrete}).

\begin{remark}[Blahut-Arimoto algorithm with particles]
\label{rem:ba_particles}
Since the optimal prior is discrete, we can maximize the mutual information directly by identifying the best set of atoms. We set the prior have the form $\pin^* = \sum_{i=1}^K K^{-1} \delta(w-w^i)$ where $\{w^1,\ldots, w^K \}$ are the $K$ atoms. We call these atoms ``particles''. Using standard back-propagation, we can then compute the gradient of the objective in~\cref{eq:ref_pin} with respect to each particle (note that each particle's gradient depends upon all other particles).
\end{remark}

\subsection{Visualizing the reference prior for deep networks}
\label{rem:visua_prior}

\begin{figure}
\centering
\includegraphics[width=0.5\linewidth]{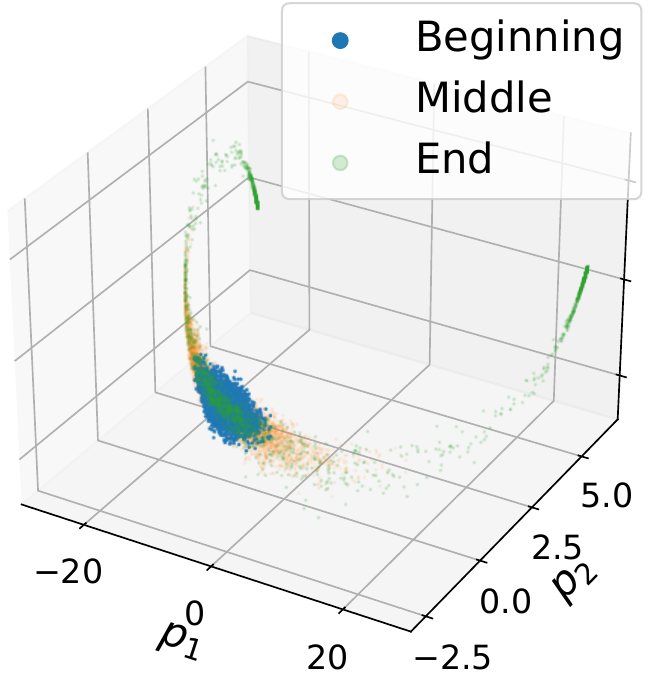}
\caption{\textbf{Reference prior (green) for binary classification on MNIST}. A three-dimensional embedding of the probability distributions of $K=3000$ atoms in the reference prior after 50,000 iterations of the BA algorithm (green) for a binary classification problem on MNIST (digits 3 vs. 5). Particles were initialized randomly (blue) and they are nearby in this embedding because at initialization, the logits of each particle are uniformly distributed. Orange shows particle locations after 5,000 iterations. As the reference prior objective in~\cref{eq:ref_pin} is optimized, the particles increasingly make more diverse predictions (orange) and towards the end (green) these particles spread apart in the prediction space.}
\label{fig:manifold_boundary}
\end{figure}

One cannot directly visualize the high-dimensional particles $w$ in $\pin^*$. But we can think of each particle $w$ as representing a probability distribution $f(w) \in \reals^{nC}$ given by
\[
    \rbr{\sqrt{p_w(y=1 \mid x_1)}, \sqrt{p_w(y=2 \mid x_1)}, \ldots, \sqrt{p_w(y=C \mid x_n)}}.
\]
and use a method for visualizing such distributions developed in~\citet{Quinn13762} that computes a principal component analysis (PCA) of such vectors $\{ f(w^1), \ldots, f(w^K)\}$ shown in~\cref{fig:manifold_boundary}. See~\cref{s:app:visualizing} for more details.

This experiment demonstrates that we can instantiate reference priors for deep networks in a scalable fashion even for a large number of particles $K$. It provides a visual understanding of how atoms of the prior are diverse models in prediction space, just like the atoms in~\cref{fig:coin_convergence_jeffreys}.

\paragraph{How to choose the number of atoms $K$ in the reference prior?}
Each particle in this paper is a deep network, so must be careful to ensure that we do not maintain an unduly large number of atoms in the prior. \citet{abbottScalingLawDiscrete2019} suggest a scaling law for $K$ in terms of the number of samples $n$, e.g., $K \sim n^{4/3}$ for a problem with two biased coins. We will instead treat $K$ as a hyper-parameter. This choice is motivated from the emergent low-dimensional structure of the green particles in~\cref{fig:manifold_boundary}; see the further analysis in in~\cref{s:expt:analysis}.
%We hypothesize that the reference prior puts probability mass on simple low-dimensional models~\citep{mattingly2018maximizing}; investigating this hypothesis is left for future work.

\begin{remark}[Variational approximations of reference priors]
\label{rem:variational_approximation}
\citet{nalisnick2017variational} maximize a lower bound on $I_\pi(w; z)$ and replace the term $p(z) = \int \dd{w} \pi(w) p(z \mid w)$ in~\cref{eq:ref_defn} by the so-called VR-max estimator $\max_w \log p(z \mid w)$ where the maximum is evaluated across a set of samples from $\pi(w)$~\citep{liEnyiDivergenceVariational2016}. They use a continuous variational family parameterized by neural networks. However, reference priors are supported on a discrete set. Using a continuous variational family, e.g., a Gaussian distribution, to approximate $\pin^*$ is computationally beneficial but it is detrimental to the primary purpose of the prior, namely to discover diverse models. This is also seen in~\cref{fig:manifold_boundary} where it would be difficult to construct a variational family whose distributions put mass mostly on the green points. We therefore do not use variational approximations.
\end{remark}

\begin{remark}[Reference prior depends upon the number of samples and its atoms are diverse models]
\label{rem:small_diverse}
\cref{eq:ref_defn} encourages the likelihood $p(\zn \mid w)$ of atoms in the reference prior to be maximally different from that of other atoms. This gives us intuition as to why the prior should have finite atoms. Consider the covering number in learning theory~\citep{bousquet2003introduction} where we endow the model space with a metric that measures disagreement between two hypotheses over $n$ samples. Smaller the number of samples $n$, smaller the covering number, and smaller the effective set of models considered. The reference prior is similar. If we only have few samples $n$, then it is not possible for the likelihood in Bayes law to distinguish between a large set of models and assign them different posterior probabilities. The prior therefore puts probability mass only on a finite set of atoms, and just like the coin-tossing experiment in~\cref{eg:bias}, these atoms have diverse outputs on the $n$ samples. This ability of the prior to select a small set of representative models is extremely useful for training deep networks with few data and it was our primary motivation.
\end{remark}

\subsection{Reference priors for semi-supervised learning}
\label{s:ssl}

Consider the situation where we are \textbf{given inputs $\xn$, their corresponding labels $\yn$ and unlabeled inputs $\xu$}. Our goal is semi-supervised learning, i.e., to use $\xu$ to build a prior $\pi^*(w)$ that selects models that can be learned using the labeled data $(\xn, \yn)$. Recall that since $\pi^*$ is a prior, it should not depend on $(\xn, \yn)$.
Just like the construction of the reference prior in~\cref{s:reference_priors}, we can maximize
\beq{
    \aed{
        I_\pi(\yn, \xn; w) & = \E_{\xn, (\yn \mid \xn, w), w \sim \pi} \sbr{\log \f{p(\yn \mid \xn, w)}{p_\pi(\yn \mid \xn)}}\\
       &= \E_{\xn, (\yn \mid \xn, w), w \sim \pi} \sbr{\log p(\yn \mid \xn, w)}
       - \E_{\xn, \yn \mid \xn} \sbr{\log p_\pi(\yn \mid \xn)}\\
        & =  \E_{\xu} \sbr{H(\yu \mid \xu)} - \E_{\xu, w \sim \pi} \sbr{H(\yu \mid \xu, w)},
        }
    \label{eq:ref_ssl}
}
where $p_\pi(\yn \mid \xn) = \int \dd{w} \pi(w) \prod_{i=1}^n p(y_i \mid x_i, w)$ and likewise for $p_\pi(\yu \mid \xu)$. The first step is simply the definition of $I_\pi$: it is the KL-divergence of the posterior after seeing $(\xn, \yn)$ with respect to the prior $\pi(w)$. The second step is the key idea and its rationale is as follows. If we know that inputs $\xu$ and $\xn$ come from the same task, then we can use samples $\xu$ to compute the expectation over $\xn$. For the same reason, we can average over outputs $\yu$ which are predicted by the network in place of the fixed labels $\yn$. Let us emphasize that both $\xu$ and $\yu$ are averaged out in the objective above. Predictions on new samples $x$ are made using the Bayesian posterior predictive distribution
\beq{
    \aed{
    p(y \mid x, \xn, \yn)
    % &= \int \dd{w} p(y \mid x, w) p(w \mid \xn, \yn)\\
    \propto \int \dd{w}  \pin^*(w) p(y \mid x, w) p(\yn \mid \xn, w).
    }
    \label{eq:bayes_posterior_predictive_distribution}
}

\paragraph{An intuitive understanding of~\cref{eq:ref_ssl}} Assume for now that we know the number of classes $C$ (although the objective is valid even if that is not the case). If our prior has $K$ particles, then the second term is the average of the per-particle entropy of the predictions. The objective encourages each particle $w_i$ to predict confidently, i.e., to have a small entropy in its output distribution $p_{w_i}(y \mid x)$. The first term is the entropy of the average predictions: $p_\pi(\yn \mid \xn)$, and it is large if particles predict different outputs $\yn$ for the same inputs $\xn$, i.e., they disagree with each other. We treat the constant $\a$ (which should be 1 in the definition of mutual information) as a hyper-parameter to allow control over this phenomenon. \textbf{The reference prior semi-supervised learning objective encourages particles to be dissimilar but confident models (not necessarily correct).}

\subsection{Reference priors for a two-stage experiment}
\label{s:two_stage}
We first develop the idea using generic random variables $\zn$. Consider a situation when we \textbf{see data in two stages, first $\zm$, and then $\zn$}. How should we select a prior, and thereby the posterior of the first stage, such that the posterior of the second stage makes maximal use of the new $n$ samples?  We can extend the idea in~\cref{s:ssl} in a natural way to address this question. We can \textbf{maximize the KL-divergence between the posterior of the second stage and the posterior after the first stage, on average, over samples $\zn$.}

Since we have access to samples $\zm$, we need not average over them, we can compute the posterior $p(w \mid \zm)$ from these samples given the prior $\pi(w)$. First, notice that $p( w, \zn \mid \zm) = p(w \mid \zmn) p(\zn \mid \zm) =  p(\zn \mid w) p(w \mid \zm) $. We can now write
\beq{
    \aed{
        \pinm^* &= \argmax_\pi I_{p(w \mid \zm)}(w; \zn)\\
        &:= \int \dd{\zn} p(\zn \mid \zm )\ \KL(p(w \mid \zmn), p(w \mid \zm ) ) \\
        % &= \int \dd{\zn} p(\zn \mid \zm ) \int \dd{w} p(w \mid \zmn) \log \f{p(w \mid \zmn)}{p(w \mid \zm)} \\
        % &= \int \dd{\zn} \dd{w} p( \zn \mid w ) p(w \mid \zm) \log \f{ p( \zn \mid w ) p(w \mid \zm) }{p(w \mid \zm) p( \zn \mid \zm )} \\
        & = \int \dd{w} p(w \mid \zm) \int \dd{\zn}p( \zn \mid w )  \log \f{ p( \zn \mid w )  }{ p( \zn \mid \zm )},
    }
}
where $p(w \mid \zm) \propto p( \zm \mid w) \pi(w)$ and $p( \zn \mid \zm ) = \int \dd{w} p( \zn \mid w) p(w \mid \zm)$. The key observation is that if the reference prior~\cref{eq:ref_pin} has a unique solution, we should have that the optimal $p(w \mid \zm) \equiv \pin^*(w)$. This leads to
\beq{
    \pinm^*(w) \propto \pin^*(w)\ p( \zm \mid w)^{-1}.
    \label{eq:ref_pinm}
}
This prior puts \emph{less} probability on regions which have high likelihood on old data $\zm$ whereby the posterior is maximally informed by the new samples $\zn$. Given knowledge of old data, the prior \emph{downweighs regions} in the weight space that could bias the posterior of the new data. We also have $\pinm^* = \pin^*$ for $m = 0$ which is consistent with~\cref{eq:ref_pin}. As $m \to \infty$, this prior ignores the part of the weight space that was ideal for $\zm$. See~\cref{s:app:two_stage_coin_tossing} for an example.

\begin{remark}[Averaging over $\zm$ in the two-stage experiment]
If we do not know the outcomes $\zm$ yet, the prior should be calculated by averaging over both $\zm, \zn$
\beq{
    \aed{
        &\pi^* = \argmax_\pi \int \dd{\zm} p(\zm) I_{p(w \mid \zm)}(w; \zn)\\
        % &:= I_\pi(w; \zmn \mid \zm)\\
        &:= I_\pi(w; \zmn) - I_\pi(w; \zm)
        = H(w \mid \zm) - H(w \mid \zmn).
    }
    \label{eq:ref_pinm_avg}
}
The encourages multiple explanations of initial data $\zm$, i.e., high $H(w \mid \zm)$, so as to let the future samples $\zn$ select the best one among these explanations, i.e., reduce the entropy $H(w \mid \zmn)$. It is interesting to note that neither is this two-stage prior equivalent to maximizing $I_\pi(w; \zmn)$, nor is it simply the optimal prior corresponding to objectives $I_\pi(w; \zm)$ or $I_\pi(w; \zn)$. Both~\cref{eq:ref_pinm,eq:ref_pinm_avg} therefore indicate that two-stage priors are useful when we have \emph{some} data \emph{a priori}, this can be either unlabeled samples from the same task, or labeled samples from some other task.
\end{remark}

\begin{remark}[A softer version of the two-stage reference prior]
The objective in~\cref{eq:ref_pinm_avg} resembles the predictive information bottleneck (IB) of~\citet{bialek2001predictability}, or its variational version in~\citet{alemi2020variational}, which seek to learn a representation, say $w$, that maximally forgets past data while remaining predictive of future data
\beq{
    \textstyle \max_{p(w \mid \zm)} I(w; \zn) - \b I(w; \zm).
    \label{eq:pib}
}
The parameter $\b$ in~\cref{eq:pib} gives this objective control over how much information from the past is retained in $w$.
We take inspiration from this and construct a variant of~\cref{eq:ref_pinm}
\beq{
        \aed{
            \pinm^{\b}(w) &\propto \pin^*(w) p( \zm \mid w)^{-\b}\quad \text{for}\ \b \in (0,1).\\
            \implies p( w \mid \zmn ) &\propto p(\zn \mid w) p(\zm \mid w)^{1-\b} \pin^*(w).
        }
        \label{eq:ref_pinm_beta}
}
We should use $\b=0$ when we expect that data from the first stage $\zm$ is similar to data $\zn$ from the second stage. This allows the posterior to \emph{benefit} from past samples. If we expect that the data are different, then $\b=1$ ignores regions in the weight space that predict well for $\zm$. This is similar to the predictive IB where a small $\b$ encourages remembering the past and $\b=1$ encourages forgetting.
%The difference between our reference prior approach and the predictive IB is that the prior in the latter need not be a reference prior.
\end{remark}

\subsection{Reference priors for transfer learning}
\label{s:transfer}

Consider the two-stage experiment where in the first stage we obtain $m$ samples $(\xms, \yms)$ from a ``source'' task $\ps$ and the second stage consists of $n$ samples $(\xnt, \ynt)$ from the ``target'' task $\pt$. Our goal is to calculate a prior $\pi(w)$ that best utilizes the target task data.
%The notation here is dense but it is necessary for precise exposition.

Bayesian inference for this problem involves first computing the posterior $p(w \mid \xms, \yms )\propto p(\yms \mid w, \xms) \pi(w)$ from the source task and then using it as a prior to compute the posterior for the target task $p(w \mid \xnt, \ynt, \xms, \yms)$. Just like~\cref{s:reference_priors}, \textbf{the key idea again is to maximize the KL-divergence between the two posteriors} $\KL \rbr{p(w \mid \xnt, \ynt, \xms, \yms),\ p(w \mid \xms, \yms)}$, but  averaged over samples $\xms$ and $\xnt$.

\paragraph{Case 1: Access to unlabeled data from the source $\xms$ and the target task $\xnt$}
We should average the KL-divergence over both the source and target predictions $\yms$ and $\ynt$ and maximize
\beq{
    \aed{
        \E_{\xms, \xnt, \yms \mid \xms, \ynt \mid \xnt}  \KL \rbr{p(w \mid \xnt, \ynt, \xms, \yms ), p( w \mid \xms, \yms)}
    }
    \label{eq:ref_pi_transfer_1}
}
over the prior $\pi$. Here $p_{\pi}( \yms \mid \xms) = \E_{w \sim \pi} p(\yms \mid \xms, w)$ and $p_{\pi}( \ynt \mid \xnt) = \E_{w \sim \pi} p(\ynt \mid \xnt, w)$, respectively. Note that averages over $\xms$ and $\xnt$ are computed using samples while averages over $\yms \mid \xms$ and $\ynt \mid \xnt$ are computed using the model's predictions.

\paragraph{Case 2: $\xms, \yms$ are fixed and known, and we have a pool of unlabeled target data $\xnt$}
Since we already know the labels for the source task, we will only average over $\xnt$ and $\ynt$ and maximize
\beq{
    \aed{
        \E_{ \xnt, \ynt \mid \xnt}  \KL \rbr{p(w \mid \xnt, \ynt, \xms, \yms), p( w \mid \xms, \yms )};
    }
    \label{eq:ref_pi_transfer_2}
}
here $p_{\pi}(\ynt \mid \xnt) = \int \dd{w} \pi(w) p(\ynt \mid \xnt, w)$.
%Note that we need not have a separate pool of unlabeled data from the target, we can simply use the samples $\xnt$ and average over the labels; this also applies to~\cref{eq:ref_ssl}.

\begin{remark}[Connecting~\cref{eq:ref_pi_transfer_1,eq:ref_pi_transfer_2} to practice]
Both objectives can be written down as
\beq{
    \pi^* = \argmax_\pi I_\pi(w; \ynt, \xnt, \xms, \yms) - I_\pi(w; \xms, \yms)
    \label{eq:ref_transfer_mi}
}
with the distinction that while in Case 1, we average over all quantities, namely $p(\xms), p(\yms), p(\xnt), p(\ynt)$ while in Case 2, we fix $\xms$ and $\yms$ to the provided data from the source task. Case 2 is what is typically called transfer learning. Case 1, where one has access to \emph{only unlabeled data} from a source task \emph{that is different from the target task} is not typically studied in practice. Like~\cref{eq:ref_pinm_beta}, we can again introduce a coefficient $\b$ on the second term in~\cref{eq:ref_transfer_mi} to handle the relatedness between source and target tasks.

% If the source and target tasks are the same, then the objective in~\cref{eq:ref_transfer_mi} is the same as that of~\cref{eq:ref_ssl}. If they are not, our theory indicates that the prior in~\cref{eq:ref_transfer_mi} should \emph{minimize} the mutual information with respect to the source task; this is similar to the argument in~\cref{s:two_stage}. Like~\cref{eq:ref_pinm_beta}, we can again introduce a coefficient $\b$ on the second term in~\cref{eq:ref_transfer_mi} to control handle the relatedness between source and target tasks.
\end{remark}

\subsection{Practical tricks for implementing reference priors}
\label{s:impl}

The reference prior objective is conceptually simple but it is difficult to implement it directly using deep networks and modern datasets. We next discuss some practical tricks that we have developed.

\paragraph{(1) Order of the reference prior $n$ versus the number of samples}
\citet{bernardo1979reference} set the order of the prior $n$ to be the same as the number of samples. We observe that both do not have to be identical and make a distinction between the two. In our expierments, we restrict the order to $n = 2,3$. Mathematically, this amounts to computing averages in~\cref{eq:ref_pin} or~\cref{eq:ref_ssl} over only sets of $n$-tuples. This significantly reduces the class of models considered in the reference prior by \emph{pretending} that there is a small number of samples available for training the task---which is useful, and also true in practice, for over-parametrized deep networks. This choice is also motivated by the low-dimensional structure in the reference prior in~\cref{fig:manifold_boundary}. Note that we are \emph{not} restricting to small order $n$ for computational reasons, i.e., computing the expectation over all classes $\yn$ in~\cref{eq:ref_ssl} can be done in a single forward pass.

\paragraph{(2) Using cross-entropy loss to bias particles towards good parts of the weight space}
The posterior~\cref{eq:bayes_posterior_predictive_distribution} suggests that we should first compute the prior, and then weight each particle by the likelihood of the labeled data. In practice, we combine these two steps into a single objective
\beq{
    \aed{
            \max_\pi  \g I_\pi(w; \yu, \xu) +  \E_{w \sim \pi}\sbr{\log p(\yn \mid \xn, w)},
           % \max_\pi   I_\pi(w; \yu, \xu) +  \E_{w \sim \pi}\sbr{\log p(\yn \mid \xn, w)},
        }
    \label{eq:ref_mi_ce}
}
where $\g$ is a hyper parameter, $\xn, \yn$ are labeled samples. \cref{eq:ref_mi_ce} allows us to directly obtain particles that both have high probability under the prior and a high likelihood. This is different from the correct Bayesian posterior (which would set $\g = 1$, we use $\g = 1/2$) but it is a trick often employed in the SSL literature. The second term restricts the search space for the particles in $\pi(w)$.

\paragraph{(3) Data augmentation}
State of the art SSL methods use heavy data augmentation, e.g., RandAugment~\citep{cubuk2020randaugment} and CTAugment~\citep{berthelot2019remixmatch}, which both have about 20 transformations. Some are weak augmentations such as mirror flips and crops while some others are strong augmentations such as color jitter. Methods such as FixMatch~\citep{sohn2020fixmatch} or MixMatch~\citep{berthelot2019mixmatch} use weak augmentations to get soft labels for predictions on strong augmentations.

We compute the entropy term $H(\yu \mid \xu, w)$ in~\cref{eq:ref_ssl} using the distribution $p_G(y \mid x, w) = \mathbb{E}_{g \sim G} [p_w(y \mid g(x), w)]$ where $G = G_1 \cup G_2$ is the set of weak ($G_1$) and strong ($G_2$) augmentations. Let $g_i \sim G_i$ be an augmentation and denote $p_{g_i} \equiv p_w(y \mid g_i(x), w)$ for $i \in \{1, 2\}$. In every mini-batch we use $p_G(y \mid x, w) \approx \t p_{g_1} + (1 - \t) p_{g_2}$ where $\tau$ is a hyper-parameter. This gives accuracy that is reasonable (about 87\% for 500 samples) but a bit lower than state of the art SSL methods. We noticed that if we use an upper bound on the entropy from Jensen's inequality
\beq{
    \scalemath{1}{
    - \E_{\xu} \int \dd{\yu}  p_G(\yu \mid \xu, w)\sbr{ \t \log p_{g_1} + (1 - \t)\log p_{g_2}}} 
    \label{eq:tau}
}
then we can close this gap in accuracy (see~\cref{tab:cifar10_ssl}). This is perhaps because the cross-entropy terms, e.g., $-p_{g_1} \log p_{g_2}$, force the predictions of the particles to be consistent across both types of augmentations, just like the objective in FixMatch or MixMatch. Our formulation is thus useful to not only understand SSL but also to tweak it to perform as well as current methods and thereby shed light on the theoretical underpinnings of their performance.

\paragraph{(4) Computing $H(\yu \mid \xu, w)$}
A number of SSL methods work by creating pseudolabels from weakly augmented data, which seems to be a key ingredient of good accuracy in our experience with these methods. We tried two heuristics to compute the entropy term $H(\yu \mid \xu, w)$ that are motivated by these papers. First, we follow FixMatch and only use unlabeled data with confident predictions to compute $H(\yu \mid \xu, w)$. A datum $x$ contributes to the objective only if $\max_y p_w(y|g_1(x), w) > 0.95$. Changing this threshold does not lead to deterioration of the accuracy as we see in~\cref{tab:app:ablation2}, so this heuristic need not be used while building the reference prior. Second, if $G_1$ is the set of weak augmentations (see previous point), methods like FixMatch and MixMatch use $\argmax_y p(y \mid g_1(x), w)$ as a pseudo-label but do not update this using the back-propagation gradient. This prevents the more reliable predictions on $G_1$ from changing. As a result, the entropy term $-\tau^2 p_{g_1} \log p_{g_1}$ is a constant in~\cref{eq:tau}. To normalize the terms coming from $\t$ in~\cref{eq:tau}, we set $\gamma$ in~\cref{eq:ref_mi_ce} to $1/(1 - \tau^2)$ instead of 1. We have also developed an argument to choose the appropriate value of $\t=1/3$ that we explain in~\cref{s:app:setup}. This second heuristic seems essential, in~\cref{tab:app:ablation2}, we obtain only 10\% accuracy without this heuristic.

%% file: experiments.tex
\section{Empirical Study}
\label{s:expt}

\subsection{Setup}

We evaluate on CIFAR-10 and CIFAR-100~\citep{krizhevsky2009learning}. For SSL, we use 50--1000 labeled samples, i.e., 5--100 samples/class and use the rest of the samples in the training set as unlabeled samples. For transfer learning, we construct 20 five-way classification tasks from CIFAR-100 and use 1000 labeled samples from the source and 100 labeled samples from the target task. All experiments use the WRN 28-2 architecture~\citep{zagoruyko2016wide}, same as in~\citet{berthelot2019mixmatch}.

For all our experiments, the reference prior is of order $n=2$ and has $K=4$ particles. We run all our methods for 200 epochs, with $\tau=1/3$ in~\cref{eq:tau} and $\alpha=0.1$ in~\cref{eq:ref_ssl}. We set $\gamma = (1 - \tau^2)^{-1}$ as discussed in~\cref{s:impl}. For inference, each particle maintains an exponential moving average (EMA) of the weights (this is common in SSL~\citep{tarvainen2017mean}). \cref{s:app:setup} provides more details.

%p_avg = gamma * weak + (1 - gamma) * strong
% L = CEloss + 1 / (1 - \gamma^2) MILoss // Have to add explanation for how this constant arises

\subsection{Semi-supervised learning}
\label{s:expt:ssl}

\paragraph{Baselines} We compare to a number of recent methods such as FixMatch~\citep{sohn2020fixmatch}, MixMatch~\citep{berthelot2019mixmatch}, DASH~\citep{xu2021dash}, SelfMatch~\citep{kim2021selfmatch}, Mean Teacher~\citep{tarvainen2017mean}, Virtual Adversarial Training~\citep{miyato2018virtual}, and Mixup~\citep{berthelot2019mixmatch}.

{
\renewcommand{\arraystretch}{1.25}
\begin{table}[htpb]
    \centering
    \Large
    \rowcolors{1}{}{black!5}
    \resizebox{0.75\linewidth}{!}{
    % \begin{tabular}{l|rrrrr}
    \begin{tabular}{llllll} % changed this because alignment looks better
    \rowcolor{white}
    \toprule
    Method &  \multicolumn{5}{c}{Samples} \\
        \rowcolor{white}
     & 50 & 100 & 250 & 500 & 1000  \\
    \midrule
    % PiModel          & - &  -        & 46.58 & 58.18 & 68.47   \\
    % PseudoLab        & - &  -        & 50.02 & 59.45 & 69.09   \\
    Mixup            & - &  -        & 52.57 & 63.86 & 74.28  \\
    VAT              & - &  -        & 63.97 & 73.89 & 81.32 \\
    Mean Teacher     & - &  -        & 52.68 & 57.99 & 82.68  \\
    MixMatch         & 64.21\tstar & 80.29\tstar & 88.91\tstar & 90.35\tstar & 92.25\tstar \\
    FixMatch (RA)  & \entry{86.19}{3.37} (40) & 90.12\tstar &  \entry{94.93}{0.65} & 93.91\tstar & 94.3\tstar \\
    FixMatch (CTA)   & \entry{88.61}{3.35} (40) & - & \entry{94.93}{0.33} & - & - \\
    DASH (RA)        & \entry{86.78}{3.75} (40) & - & \entry{95.44}{0.13} & - & - \\
    DASH (CTA)        & \entry{90.84}{4.31} (40) & - & \entry{95.22}{0.12} & - & - \\
    SelfMatch        & \entry{93.19}{1.08} (40)  & - & \entry{95.13}{0.26} & - & - \\
    FlexMatch        & \entry{95.03}{0.06} (40)  & - & \entry{95.02}{0.09} & - & - \\
    \midrule
    Deep Reference Prior & \entry{85.45}{2.12} & \entry{88.53}{0.67} & \entry{92.13}{0.39} & \entry{92.94}{0.22} & \entry{93.48}{0.24} \\
    \bottomrule
    \end{tabular}
    }
\caption{
\textbf{Classification accuracy of different semi-supervised learning methods on CIFAR-10.} \textbf{Note:} RA and CTA in the methods column indicate that RandAugment or CTAugment were used for augmentations. Entries with * were evaluated by us using open-source implementations from the original authors for 256 epochs. All other entries are from original papers. Entries with ``(40)'' indicate that 40 labeled samples were used instead of 50.
}
\label{tab:cifar10_ssl}
\end{table}
}

\cref{tab:cifar10_ssl} compares the accuracy of different SSL methods on CIFAR-10. We find that the reference prior approach is competitive with a number of existing methods, e.g., it is remarkably close to FixMatch on all sample sizes (notice the error bars). There is a gap in accuracy at small sample sizes (40--50) when compared to recent methods. It is important to note that these recent methods employ a number of additional tricks, e.g., FlexMatch implements curriculum learning on top of FixMatch, DASH and FlexMatch use different thresholding for weak augmentations (this increases their accuracy by 2-5\%), SelfMatch has higher accuracies because of a self-supervised pretraining stage, FixMatch (CTA) outperforms its RA variant by 1.5\% which indicates CTA augmentation is beneficial (we used RA). It is also extremely expensive to train SSL algorithms for 1000 epochs (all methods in~\cref{tab:cifar10_ssl} do so), we trained for 200 epochs.

This experiment shows that our approach to SSL can obtain results that are competitive to sophisticated empirical methods without being explicitly formulated to enforce properties like label consistency with respect to augmentations. This also indicates that reference priors could be a good way to explain the performance of these existing methods, which is one of our goals in this paper.

\subsection{Transfer learning}
\label{s:expt:transfer}

Just like we did in~\cref{s:impl} for SSL, we instantiate~\cref{eq:ref_pinm_beta} and~\cref{eq:ref_pi_transfer_2}, by combining prior selection, pretraining on the source task and likelihood of the target task, into one objective,
\beq{
    \aed{
    \textstyle \g I_\pi(w; \yu_t, \xu_t) &+ \E_{w \sim \pi} \sbr{\log p(w, \ynt \mid \xnt)}
    + (1 - \b) \E_{w \sim \pi} \sbr{\log p(w, \yms \mid \xms)},
    }
    \label{eq:transfer_expt_obj}
}
where $\g = 1/2$ and $\b=1/2$ are hyper-parameters, $(\xms, \yms)$ are labeled data from the source task ($m=1000$), $(\xnt, \ynt)$ are labeled data from the target task ($n=100$) and $\xu_t$ are unlabeled samples from the target task (all other samples).

\begin{figure}[!h]
\centering
\includegraphics[width=0.3\linewidth]{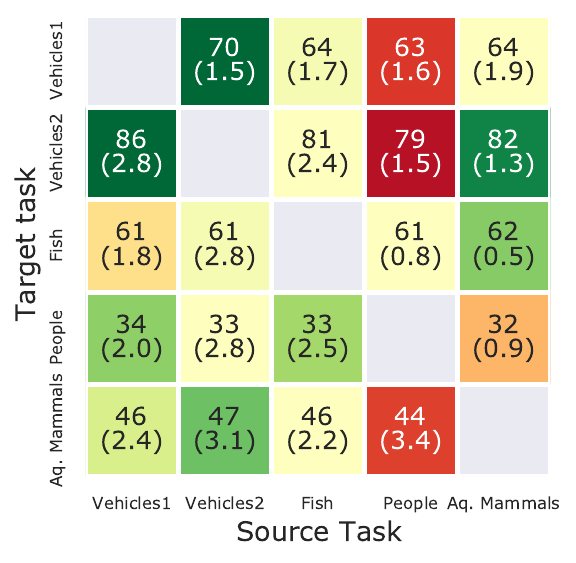}
\includegraphics[width=0.3\linewidth]{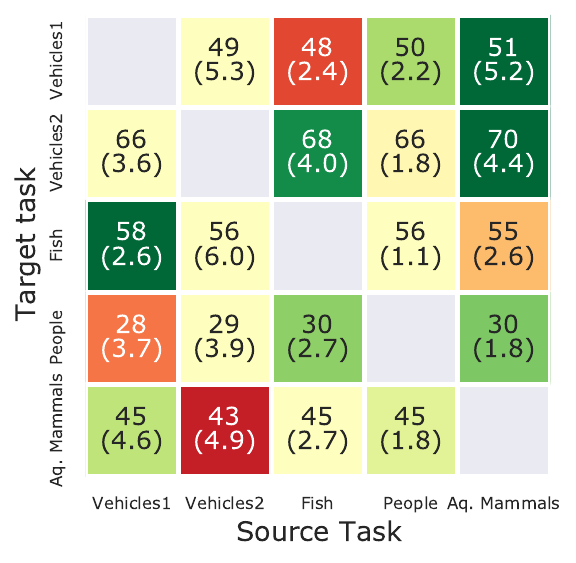}
{
\renewcommand{\arraystretch}{1.2}
\begin{table}[H]
    \centering
    \LARGE
    \rowcolors{1}{}{black!5}
    \resizebox{0.75\linewidth}{!}{
    \begin{tabular}{l|rrrrr}
    \rowcolor{white}
    \toprule
    \rowcolor{white}
    Method \qquad \qquad Task ($\rightarrow$)
    & Vehicles-1 & Vehicles-2 & Fish & People & Aq. Mammals  \\
    \midrule
    Supervised Learning     & 42.2 & 63.2 & 56.8 & 31.0 & 42.6   \\
    Deep Reference Prior (SSL)   & 63.6 & 75.2 & 54.6 & 34.0 & 47.4 \\     \bottomrule 
    \end{tabular}
    }
\end{table}
}
\caption{\textbf{Top: Accuracy (\%) of deep reference priors (left) and fine-tuning (right) for transfer learning tasks in CIFAR-100}. Cells are colored red/green relative to the median accuracy of each row. Darker shades of green indicate that the source task is more suitable for transfer. For example, Vehicles-1 as source is the best for all tasks according to the reference prior (left) (which is optimal in theory) but fine-tuning cannot replicate this. The accuracy of cells in the left panel is better than the corresponding cells on the right, e.g., the gap in accuracy is 34.8\% for Vehicles 2 $\rightarrow$ Vehicles 1.
\textbf{Bottom: Accuracy (\%) of supervised learning and SSL for all 5 tasks}. Each number here should be compared to the corresponding row of the matrices in the top panel, e.g., Vehicles 2 has 86\% accuracy when transferred from Vehicles 1 using our transfer method (left), it has 66\% accuracy from fine-tuning (right), while the same task achieves 63.2\% accuracy when trained by itself using supervised learning (table first row) and 75.2\% accuracy when trained using unlabeled target data (table second row). Therefore the reference prior-based transfer objective can leverage both labeled source data as well as unlabeled target data. This pattern is consistent for all tasks.
}
\label{fig:matrix}
\end{figure}

\paragraph{Baselines} We use three methods: (a) fine-tuning, which is a very effective strategy for transfer learning~\citep{dhillon2019a,kolesnikovbigtransferbit2020} but it cannot use unlabeled target data, (b) using only labeled target data (this is standard supervised learning), and (c) using only labeled and unlabeled target data without any source data (this is simply SSL, or $\b=1$ in~\cref{eq:transfer_expt_obj}). \cref{fig:matrix} compares the performance for pairwise transfer across 5 tasks from CIFAR-100. Our reference prior objective in~\cref{eq:transfer_expt_obj} obtains much better accuracy than fine-tuning which indicates that it leverages the unlabeled target data effectively. For each task, the accuracy is much better than both standard supervised learning and semi-supervised learning using our own reference prior approach~\cref{eq:ref_mi_ce}; both of these indicate that the labeled source data is being used effectively in~\cref{eq:transfer_expt_obj}.

\subsection{Ablation and analysis}
\label{s:expt:analysis}

This section presents ablation and analysis experiments for SSL on CIFAR-10 with 1000 labeled samples. We study the reference prior for different settings (i) varying the order $n$ of the prior, (ii) varying the number of particles in the BA algorithm ($K$), (iii) exponential moving averaging of the weights for each particle. We also study the two entropy terms in the reference prior objective individually.

We use a reference prior of order $n=2$ in all our experiments. We see in~\cref{tab:app:order} that \textbf{changing the order of the prior} leads to marginal (about 1\%) changes in the accuracy.

\begin{table}[htpb]
    \centering
    \rowcolors{1}{}{black!5}
    \resizebox{0.7\linewidth}{!}{
    \begin{tabular}{l|rrrr}
    \rowcolor{white}
    \toprule
    Method \qquad \qquad Order ($\rightarrow$)
    & 2 & 3 & 4 & 5   \\
    \midrule
    Deep Reference Prior ($K=4$)  & 91.76 & 90.53 & 91.51 & 91.36 \\     \bottomrule 
    \end{tabular}
    }
\caption{The order of the reference prior has a minimal impact on the accuracy.}
\label{tab:app:order}
\end{table}

\begin{table}[htpb]
    \centering
    \rowcolors{1}{}{black!5}
    \resizebox{0.7\linewidth}{!}{
    \begin{tabular}{l|rrrr}
    \rowcolor{white}
    \toprule
    \rowcolor{white}
    Method \qquad \qquad \#Particles ($\rightarrow$)
    & 2 & 4 & 8 & 16   \\
    \midrule
    Deep Reference Prior ($n=2$)
    & 91.3 & 91.76 & 89.79 & 90.72 \\     \bottomrule 
    \end{tabular}
    }
\caption{Number of particles has a minimal impact on accuracy.}
\label{tab:app:particles}
\end{table}

We next \textbf{vary the number of particles} in the prior in~\cref{tab:app:particles} and find that the accuracy is relatively consistent when the number of particles varies from $K=2$ to $K=16$. This seems surprising because a reference prior ideally should have an infinite number of atoms, when it approximates Jeffreys prior. We should not a priori expect $K=2$ particles to be sufficient to span the prediction space of deep networks. But our experiment in ~\cref{fig:manifold_boundary} provides insight into this phenomenon. It shows that the manifold of diverse predictions is low-dimensional. Particles of the reference prior only need to span these few dimension and we can fruitfully implement our approach using very few particles.

\paragraph{Effect of exponential moving averaging (EMA)}
We use EMA on the weights of each particle (independently). \cref{tab:app:ema} analyzes the impact of EMA. As noticed in other semi-supervised learning works~\citep{berthelot2019mixmatch,sohn2020fixmatch}, EMA improves the accuracy by 2-3\% regardless of the number of labeled samples used.

\begin{table}[htpb]
\renewcommand{\arraystretch}{1.2}
    \centering
    \rowcolors{1}{}{black!5}
    \resizebox{0.9\linewidth}{!}{
    \begin{tabular}{l rrrrr}
    \rowcolor{white}
    \toprule
    \rowcolor{white}
    Method \qquad \#Samples ($\rightarrow$)
    & 50 & 100 & 250 & 500 & 1000   \\
    \midrule
    EMA & \entry{85.45}{2.12} & \entry{88.53}{0.67} & \entry{92.13}{0.39} & \entry{92.94}{0.22} & \entry{93.48}{0.24} \\
    No EMA & \entry{82.36}{2.13} & \entry{85.64}{0.43} & \entry{89.75}{0.36} & \entry{90.06}{1.71} & \entry{91.57}{0.25} \\
    \bottomrule 
    \end{tabular}
    }
\caption{Using EMA for weights of each particle is beneficial and improves accuracy by 2-3\%.}
\label{tab:app:ema}
\end{table}

\begin{figure}[htpb]
\centering
\includegraphics[width=0.495\linewidth]{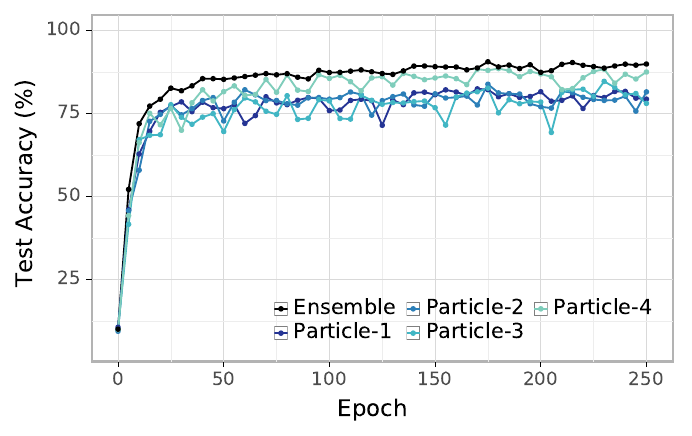}
\includegraphics[width=0.495\linewidth]{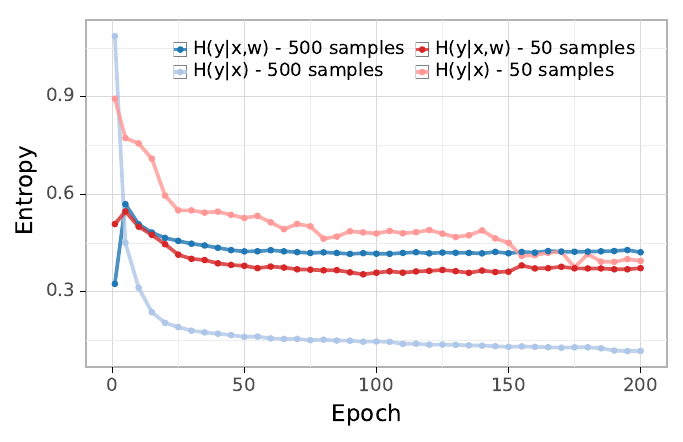}
\caption{\textbf{(Left)} Accuracy of individual particles in the prior during training (250 labeled samples). The individual particles have diverse predictions due to the entropy term $H(\yn \mid \xn)$, the accuracy of the ensemble is larger than the accuracy of any single particle.
\textbf{(Right)} Evolution of entropy terms $H(\yu \mid \xu, w)$ and $H(\yu \mid \xu)$ for two cases (500 labeled samples and 50 labeled samples). While $H(\yu \mid \xu)$ is expected to be larger than $H(\yu \mid \xu, w)$ in~\cref{eq:ref_ssl} since KL-divergence is non-negative, this is not always the case since we approximate $H(\yu \mid \xu, w)$ by an upper-bound obtained from Jensen's inequality for data augmentation as discussed in~\cref{s:impl}.
}
\label{fig:particle_accuracy_entropy}
\end{figure}

\paragraph{The two entropy terms in the reference prior objective}
\cref{fig:particle_accuracy_entropy} (left) shows how, because of the entropy term $H(\yu \mid \xu)$, the accuracy of particles is quite different during training. Particles have different predictive abilities ( 7\% range in test error) but the Bayesian posterior predictive distribution has a higher accuracy than any of them. \cref{fig:particle_accuracy_entropy} (right) tracks the two entropy terms in the objective. For large number of labeled data (500, blue) the entropy $H(\yu \mid \xu)$ which should always be higher than $H(\yu \mid \xu, w)$ in~\cref{eq:ref_ssl} is lower (this is not the case for 50 samples, red). This is likely a result of the cross-entropy term in the modified objective in~\cref{eq:ref_mi_ce} which narrows the search space of the particles. This experiment is an important insight into the working of existing semi-supervised learning methods as well, all of which also have a similar cross-entropy objective in their formulation. It points to the fact that at large sample-sizes, the cross-entropy loss and not the semi-supervised learning objective could dominate the training procedure.

%% file: related_work.tex
% !TEX root = ./ICML_main.tex

\section{Related Work and Discussion}
\label{s:related}

\paragraph{Reference priors in Bayesian statistics}
We build upon the theory of reference priors which was developed in the objective Bayesian statistics literature~\citet{bernardo1979reference,berger1988priors,berger2009formal}. The main idea used in our work is that non-asymptotic reference priors allow us to exploit the finite samples from the task in a fundamentally different way than classical Bayesian inference. If the number of samples from the task available to the learner is finite, then the prior should also select only a finite number of models. Reference priors are not common in the machine learning literature. A notable exception is~\citet{nalisnick2017variational} who optimize a variational lower bound and demonstrate results on small-scale problems. The main technical distinction of our work is that we explicitly use the discrete prior instead of a variational approximation.

\paragraph{Information theory}
Discreteness is seen in many problems with an information-theoretic formulation, e.g., capacity of a Gaussian channel under an amplitude constraint~\citep{smith1971information}, neural representations in the brain~\citet{laughlin1981simple}, and biological systems~\citep{mayer2015well}.  \citep{mattingly2018maximizing,abbottScalingLawDiscrete2019} have developed these ideas to study how reference priors select ``simple models'' which lie on certain low-dimensional ``edges'' of the model space. We believe that the methods developed in our paper are effective \emph{because} of this phenomenon. Our choice of using a small order $n$ for the prior is directly motivated by their examples.

\paragraph{Semi-supervised learning} Our formulation sheds light on the working of current SSL methods. For example, the reference prior can automatically enforce consistency regularization of predictions across augmentations~\citep{tarvainen2017mean, berthelot2019mixmatch}, as we discuss in~\cref{s:impl}. Similarly, minimizing the entropy of predictions on unlabeled data, either explicitly~\citep{grandvalet2005semi, miyato2018virtual} or using pseudo-labeling methods~\citep{lee2013pseudo, sajjadi2016mutual}, is another popular technique. This is automatically achieved by the objective in~\cref{eq:ref_ssl}. Disagreement-based methods~\citep{zhou2010semi} employ multiple models and use confident models to soft-annotate unlabeled samples for others. Disagreements in our formulation are encouraged by the entropy $H(\yn \mid \xn)$ in~\cref{eq:ref_ssl}. If $p(\yn \mid \xn)$ is uniform, which is encouraged by the reference prior objective, particles disagree strongly with each other.

\paragraph{Transfer learning} is a key component of a large number of applications today, e.g,~\citep{devlin2019bert,kolesnikovbigtransferbit2020} but a central question that remains unanswered is how one should pretrain a model if the eventual goal is to transfer to a target task. There have been some attempts at addressing this via the Information Bottleneck, e.g.,~\citet{gao2020free}. This question becomes particularly challenging when transferring across domains, or for small sample sizes~\citep{davatzikos2019machine}. Reference priors are uniquely suited to tackle this question: our two-stage experiment in~\cref{s:two_stage} is the \emph{optimal} way pretain on the source task. As our experiments show, this is better than fine-tuning in the low-sample regime \cref{s:expt:transfer}.

%% file: appendix.tex
% !TEX root = ./ICML_main.tex

\renewcommand{\theequation}{S-\arabic{equation}}
\renewcommand{\thefigure}{S-\arabic{figure}}
\renewcommand{\thetable}{S-\arabic{table}}

\newpage
\appendix
\onecolumn

\section{Details of the experimental setup}
\label{s:app:setup}

\paragraph{Architecture} For experiments on CIFAR-10 and CIFAR-100 (\cref{s:expt}), we consider a modified version of the Wide-Resnet 28-2 architecture ~\citep{zagoruyko2016wide}, which is identical to the one used in \citet{berthelot2019mixmatch}. This architecture differs from the standard Wide-Resnet architecture in a few important aspects. The modified architecture has Leaky-ReLU with slope 0.1 (as opposed to ReLU), no activations or batch normalization before any layer with a residual connection, and a momentum of 0.001 for batch-normalization running mean and standard-deviation (as opposed to 0.1, in other words these statistics are made to change very slowly). We observed that the change to batch-normalization momentum has a very large effect on the accuracy of semi-supervised learning.

For experiments on MNIST (\cref{s:app:mnist}), we use a fully-connected network with 1 hidden layer of size 32. We use the hardtanh activation in place of ReLU for this experiment; this is because maximizing the mutual information has the effect of increasing the magnitude of the activations for ReLU networks. One may use weight decay to control the scale of the weights and thereby that of the activations but in an effort to implement the reference prior exactly, we did not use weight decay in this model. Note that the nonlinearities for the CIFAR models are ReLUs.

\paragraph{Datasets} For semi-supervised learning, we consider the CIFAR-10 dataset with the number of labeled samples varying from 50--1000 (i.e., 5--100 labeled samples per class). Semi-supervised learning experiments use all samples that are not a part in the labeled set, as unlabeled samples.

For transfer learning, we construct two tasks from MNIST (task one is a 5-way classification task for digits 0--4, and task two is another 5-way classification task for digits 5--9). For this experiment, we use labeled source data but do not use any labeled target data. This makes our approach using a reference prior similar to a purely unsupervised method.

The CIFAR-100 dataset is also utilized in the transfer learning setup (\cref{s:expt:transfer}). We consider five 5-way classification tasks from CIFAR-100 constructed using the super-classes. The five tasks considered are Vehicles-1, Vehicles-2, Fish, People and Aquatic Mammals. The selection of these tasks were motivated from the fact that some pairs of tasks are known to positively impact each other (Vehicles-1, Vehicles-2), while other pairs are known to be detrimental to each other (Vehicles-2, People); see the experiments in~\citet{rameshModelZooGrowing2022}.

\paragraph{Optimization}
SGD with Nesterov momentum on a Cosine-annealed learning rate schedule with warmup was used in our experiments on CIFAR-10 and CIFAR-100. The initial learning rate was set to $0.03 \times K$ where $K$ denotes the number of particles. The scaling factor of $K$ exists to counteract the normalization constant in the objective from averaging across all particles. The momentum coefficient for SGD was set to 0.9 and weight decay to $5K^{-1}\times 10^{-4}$. Mixed-precision (32-bit weights, 16-bit gradients) was used to expedite training. Training was performed for 200 epochs unless specified otherwise. 

Experiments on MNIST also used SGD for computing the reference prior. SGD was used with a constant learning rate of 0.001 with Nesterov's acceleration, momentum coefficient of 0.9 and weight decay of $10^{-5}$.

\paragraph{Definition of a single Epoch} Note that since we iterate over the unlabeled and labeled data (each with different number of samples), the notion of what is an epoch needs to be defined differently. In our work, one epoch refers to 1024 weight updates, where each weight update is calculated using a batch-size of 64 for the labeled data of batch size 64, and a batch-size of 448 for the unlabeled data.

\paragraph{Exponential Moving Average (EMA)}
In all CIFAR-10 and CIFAR-100 experiments, we also implement the Exponential Moving Average (EMA)~\citep{tarvainen2017mean}. In each step, the EMA model is updated such that the new weights are the weighted average of the old EMA model weights, and the latest trained model weights. The weights for averaging used in our work (and most other methods) are 0.999 and 0.001 respectively. Note that EMA only affects the particle when it is used for testing, it does not affect how weight updates are calculated during training. We exclude batch-normalization running mean and variance estimates in EMA.

\paragraph{Data Augmentations}
We use random-horizontal flips and random-pad-crop (padding of 4 pixels on each side) as weak augmentations for the CIFAR-10 and CIFAR-100 datasets. For SSL experiments on CIFAR-10, we use RandAugment~\citep{cubuk2020randaugment} for strong augmentations. No data augmentations were used for MNIST.

\paragraph{Picking the value of $\t$ in~\cref{eq:tau}}
Let $G_1$ and $G_2$ be the sets of weak and strong augmentations respectively. For $g_1 \sim G_1$ and $g_2 \sim G_2$, let us write down the upper bound in~\cref{eq:tau} from Jensen's inequality in detail
\[
 \E_{\xu}  \int \dd{ \yu}   \sbr{- \t^2 p_{g_1} \log p_{g_1}    - \t( 1- \t) p_{g_2}  \log p_{g_1}  - (1-\t)\t p_{g_1} \log p_{g_2} - (1 - \t )^2 p_{g_2} \log p_{g_2} }.
\]
The upper bound is thus a weighted sum of the entropy terms $-p_{g_1} \log p_{g_1}, -p_{g_2} \log p_{g_2}$, and cross entropy terms $-p_{g_2} \log p_{g_1}, -p_{g_1} \log p_{g_2}$. If we were to pick $\t = 1/2$ like FixMatch, then since $(1 - \t )^2 + \t^2 = 2 \t (1- \t)$ for $\t=1/2$, the entropy and cross entropy terms will contribute equally to the loss function. However in practice, since we do not update $p_{g_1}$ using the back-propagation gradient to protect the predictions from deteriorating on the weakly augmented images, one of the entropy terms $- p_{g_1} \log p_{g_1}$ is dropped. In such a situation, to ensure that cross entropy and entropy terms provide an equal contribution to the gradient, we would like $(1- \t)^2 = 2 \t (1- \t)$ which gives $\t = 1/3$.

\section{Overview of the Implementation}

We provide an overview of the implementation of deep reference priors.\\
For more details see \href{https://github.com/rahul13ramesh/deep_reference_priors}{https://github.com/rahul13ramesh/deep\_reference\_priors}.

Let a mini-batch from the labeled dataset be denoted by  $\{(x_i, y_i)\}_{i=1}^b$ and a mini-batch from the unlabeled dataset be denoted by $\{(x_{i0}^u, x_{i1}^u, \cdots, x_{in}^u))\}_{i=1}^{b_u} $ where $n$ is the order of the reference prior. Note the distinction in the two mini-batches, i.e. the unlabeled mini-batch consists of a set of n-tuples unlike the labeled mini-batch. 
Let $g_1$ and $g_2$ be functions that perform weak and strong augmentations respectively. The reference prior objective is used to train $K$ particles $\{ p_{w_k}\}_{k=1}^K$.

For the sample $\xu$, we compute $p(y \mid \xu, w_k)$ as follows:
\[ p(y \mid \xu, w_k) = \tau p(y \mid g_1(\xu), w_k) + (1 - \tau) p(y \mid g_2(\xu), w_k) \] 
The reference prior loss ,requires us to compute the terms
\begin{align*}
    \mathbb{E}_{w \sim \pi} \left[ H(\yu_i \mid \xu_i, w) \right]
    &= \sum_{k=1}^K \pi(w_k) \sum_{y \in \mathcal{Y}^n} \left( - p(y \mid \xu_i, w_k) \log(p(y \mid \xu_i, w_k)) \right) \\
    &= \sum_{k=1}^K \pi(w_k) \sum_{y \in \mathcal{Y}^n} \left( - \prod_{j=1}^n p(y \mid \xu_{ij}, w_k) \right)    \log \left( \prod_{j=1}^n p(y \mid \xu_{ij}, w_k) \right)  \\
    &= \sum_{k=1}^K \pi(w_k) \sum_{j=1}^n \sum_{y \in \mathcal{Y}} - p(y \mid \xu_{ij}, w_k) \log \left( p(y \mid \xu_{ij}, w_k) \right) \\
    &\leq \sum_{k=1}^K \pi(w_k) \sum_{j=1}^n \sum_{y \in \mathcal{Y}} - p(y \mid \xu_{ij}, w_k) \left[ \tau \log p(y \mid g_1(\xu_{ij}), w_k) + (1 - \tau) \log p(y \mid g_2(\xu_{ij}), w_k) \right],
\end{align*}
and 
\begin{align*}
    H(\yu_i \mid \xu_i)
    &= \sum_{y^n \in \mathcal{Y}^n} - p(y^n \mid \xu_i) \log(p(y^n \mid \xu_i)) \\
    &= \sum_{y^n \in \mathcal{Y}^n} - \left( \sum_{k=1}^K \pi(w_k) p(y^n \mid \xu_i, w_k) \right)
    \log \left( \sum_{k=1}^K \pi(w_k) p(y^n \mid \xu_i, w_k)  \right).
\end{align*}
In our implementation, we set $\pi(w_k) = \frac{1}{K}$. We observed no improvement in accuracy if the elements of $\pi$ were trainable weights.

\begin{figure}
\begin{framed}
\textbf{Input} data consists of a mini-batch of labeled data $\cbr{(x_i, y_i)}_{i=1}^b$ and unlabeled data $\cbr{x_{i0}^u,  x_{i1}^u, \cdots, x_{in}^u)}_{i=1}^{b_u}$ and a user-determined order $n$.

\textbf{Trainable weights} are the weights of the $K$ neural networks (also called particles) $\cbr{p_{w_k}}_{k=1}^K$.

Define
\[
    \aed{
        f(x, y, w) &= \tau p_w(y \mid g_1(x)) + (1 - \tau) p_w(y \mid g_2(x)),\\
        f_{\log}(x, y, w) &= \tau \log p_w(y \mid g_1(x)) + (1 - \tau) \log p_w(y \mid g_2(x)).
        % g(x, y, w) &= \prod_{i=1}^n f(x_i, y, w).
    }
\]

Compute the two entropy terms as
\[
    \aed{
        h_{yw} &= -\f{1}{b_u} \sum_{i=1}^{b_u} \sum_{k=1}^K \f{1}{K} \sum_{j=1}^n \sum_{y \in \mathcal{Y}} f(x_{ij}^u, y, w_k) f_{\log}(x_{ij}^u, y, w_k),\\
        h_{y} &= -\f{1}{b_u} \sum_{i=1}^{b_u} \sum_{y^n \in \mathcal{Y}^n} \rbr{\f{1}{K} \sum_{k=1}^K \prod_{j=1}^n f(\xu_{ij}, y_j^n, w)} \log \rbr{\f{1}{K} \sum_{k=1}^K \prod_{j=1}^n f(\xu_{ij}, y_j^n, w)}.
    }
\]

Compute the loss $\ell$ as
\[
    \aed{
        \ell_u &=  \alpha h_{y} - h_{yw}\\
        \ell_x &= -\f{1}{b K} \sum_{i=1}^b \sum_{k=1}^K \log (p_{w_k}(y_i \mid x_i))\\
        \ell &= \ell_x - \rbr{\f{1}{1 - \tau^2}} \ell_u.
    }
\]
\end{framed}
\caption{The pseudo-code to compute the loss of one mini-batch of data while computing the reference prior.}
\end{figure}

\section{Visualizing the reference prior}
\label{s:app:visualizing}

We can think of each particle $w$ as representing a probability distribution
\[
    \reals^{nC} \ni f(w) = \rbr{\sqrt{p_w(y=1 \mid x_1)}, \sqrt{p_w(y=2 \mid x_1)}, \ldots, \sqrt{p_w(y=C \mid x_n)}}.
\]
and use a method for visualizing such distributions developed in~\citet{Quinn13762} that computes a principal component analysis (PCA) of such vectors $\{ f(w^1), \ldots, f(w^K)\}$. This method computes an isometric embedding of the space of probability distributions. The rationale behind the choice of $f(w)$ is that for two weight vectors $w, w'$, the Euclidean distance between $f(w)$ and $f(w')$ is the Hellinger divergence between the respective probability distributions,
\beqs{
    \aed{
    \norm{f(w) - f(w')}^2 &= \f{1}{2 n} \sum_{i=1}^{n} \sum_{k=1}^{C} \rbr{ \sqrt{p_w( y = k \mid x_i )} - \sqrt{p_{w'}( y = k \mid x_i )}}^2 \\
    & = \f{1}{n} \sum_{i=1}^{n} d_H^2\rbr{p_w( \cdot \mid x_i ), p_{w'}( \cdot \mid x_i )},
    }
}
where
\[
    d^2_H(P, Q) = \f{1}{2} \int \rbr{\sqrt{\dd{P}} - \sqrt{\dd{Q}}}^2
\]
is the Hellinger distance. In other words, the prediction vector $f(w)$ maps the weights $w$ into a $(n C)-$dimensional space. The Euclidean metric in this space corresponds to the Hellinger distance in the space of probability distributions. We can therefore compute the principal component analysis (PCA) of these vectors and project the vectors $f(w)$ into lower-dimensions to visualize them, as done in~\cref{fig:manifold_boundary}.

% \begin{figure}[htpb]
% \centering
% \includegraphics[width=0.4\linewidth]{fig/Manifold_boundary_rasterized.pdf}
% \caption{We compute and visualize the reference prior for a binary classification problem on MNIST (digits 3 vs. 5). We randomly initialize $K=3000$ particles (blue) and then maximize the mutual information in~\cref{eq:ref_ssl}. For each atom $w^k$ of our prior, $k= 1,2,\ldots,3000$, we evaluate its prediction vector $f(w^k)$. By using PCA, we embed all $K$ prediction vectors into a 3-dimensional Euclidean space spanned by the top three principle components $\vec{p_1}$, $\vec{p_2}$ and $\vec{p_3}$. The different colors show the prior at initialization (blue), after 5,000 iterations (orange) and towards the end of the computation after 50,000 iterations (green).
% }
% \label{fig:app:manifold_boundary}
% \end{figure}

% At the beginning of optimization, all atoms $w^k$ make similar predictions (blue points in~\cref{fig:manifold_boundary} are near each other). This is not surprising because the atoms are initialized randomly and each $w^k$ has a uniform distribution as its output for all samples. As the prior is optimized to maximize the mutual information by the Blahut-Arimoto algorithm, its atoms increasingly make more diverse predictions (orange). Towards the end (green), predictions of these atoms seem to form a low-dimension manifold. We hypothesize that these green particles form a ``boundary'' of the possible prediction vectors made by a deep network and consist of simple, low-dimensional models~\citep{mattingly2018maximizing}.

\section{Additional Experiments} 
\label{s:app:expt}

\subsection{Unsupervised transfer learning on MNIST}
\label{s:app:mnist}

For the following experiments on MNIST, the reference prior is of order $n= 2$ and has $K= 50$ particles. We run our methods for $1024$ epochs.

We first compare deep reference priors with fine-tuning for transfer learning. The parameter $\b$ controls the degree to which the posterior~\cref{eq:ref_pinm_beta} is influenced by the target data. If we have $\beta=1$, then the posterior is maximally influenced by target data after being pretrained on the source data. We instantiate~\cref{eq:ref_pinm_beta}, by combining prior selection, pretraining on the source task into one objective,
\beq{
\aed{
 \textstyle \max_\pi \g I_\pi(w; \yu, \xu) + (1 - \b) \E_{w \sim \pi}\log p(w; y^s \mid x^s),
}
\label{eq: mnist_transfer}
}
where $\g$ and $\b$ are hyper-parameters. Solving~\cref{eq: mnist_transfer} requires no knowledge from target data labels, therefore the setting here is pure unsupervised clustering for target task dataset. We compare this objective to fine-tuning which adapts a model trained on labeled source to the labeled target data. In this experiment, all samples from the source task (about 30,000 images across 5 classes) were used for both the reference prior and fine-tuning.

\begin{table}[H]
\renewcommand{\arraystretch}{1.25}
    \centering
    \rowcolors{1}{}{black!5}
    \resizebox{0.65\linewidth}{!}{
    \begin{tabular}{l rrrrr}
    \rowcolor{white}
    \toprule
    \rowcolor{white}
    Method \qquad \# Labeled target data ($\rightarrow$)
    &0 & 50 & 100 & 250 & 500   \\
    \midrule
    \textbf{Source (0--4) to Target (5--9)}\\
    Fine-Tuning     & - & 71.1 & 78.8 & 86.6    & 93.0\\
    Deep Reference Prior Unsupervised Transfer   & 87.4 & - & - & -& -  \\
    \midrule
    \textbf{Source (5--9) to Target(0--4)}\\
    Fine-Tuning     & - & 90.2 & 92.4 & 94.7   & 96.2\\
    Deep Reference Prior Unsupervised Transfer   & 95.2 & - & - & -& -  \\     \bottomrule 
    \end{tabular}
    }
\caption{
\textbf{Accuracy (\%) of unsupervised reference-prior based transfer} (digits 0--4) to the target task (digits 5--9). We see that transfer using source and unlabeled target data using the reference prior performs as well as fine tuning with labeled source data and 250 labeled target data. Even if MNIST is a simple dataset, this is a remarkable demonstration of how effective the reference prior is at making use of both the labeled source data and unlabeled target data.
}
\label{tab:mnist_transfer1}
\end{table}

\subsection{More ablation studies}

\cref{s:impl} describes a few implementation tricks that we employ when computing $H(\yu \mid \xu, w)$.  The unlabeled samples consist of both weak and strong augmentations of the same image $x$ which we denote by $g_1(x)$ and $g_2(x)$ and we define $p_{g_i} \equiv p_w(y|g_i(x), w)$. The objective can be upper-bounded using Jensen's inequality as follows
\[
    \aed{
    H(\yu \mid \xu, w)
    &= - \E_{\xu} \int \dd{\yu}  p_G(\yu \mid \xu, w)\sbr{
    \log( p_G(\yu \mid \xu, w))} \\
    &= -\E_{\xu} \int \dd{\yu}  p_G(\yu \mid \xu, w)\sbr{
    \log( \tau p_{g_1} + (1 - \tau) p_{g_2} )} \\
    &\leq -\E_{\xu} \int \dd{\yu}  p_G(\yu \mid \xu, w)\sbr{
    \tau \log p_{g_1} + (1 - \tau) \log p_{g_2}}
    }
\]

The first trick is to use the above bound from Jensen's inequality to compute $H(\yu \mid \xu, w)$. The second trick we employ is to not update $p(y \mid g_1(x), w)$ with back-propagation gradients. \cref{tab:app:ablation2} shows that both these tricks are needed to achieve good accuracy. 

The third trick is to include $x$ in the loss only if $\max p_w(y \mid g_1(x), w) > 0.95$ -- an implementation detail also employed in \cite{sohn2020fixmatch}. \cref{tab:app:ablation2} shows that this has very little impact on accuracy. 

\begin{table}[H]
\renewcommand{\arraystretch}{1.25}
    \centering
    \rowcolors{1}{}{black!5}
    \resizebox{0.6\linewidth}{!}{
    \begin{tabular}{l r}
    \rowcolor{white}
    \toprule
    \rowcolor{white}
    Implementation trick & Accuracy (\%)\\
    \midrule
    Deep reference priors (All 3 tricks) & 92.13 \\
    No stop gradient to $p_{g_1}$  & 10\\
    No Jensen's inequality & 86.55\\
    No masking using probability threshold & 92.35 \\       \bottomrule
    \end{tabular}
    }
\caption{
We perform an ablation study over the three implementation tricks considered in~\cref{s:impl} and compute the accuracy after removing each one of the tricks. The accuracy (\%) is computed for 250 labeled samples, with 4 particles and using order 2.
}
\label{tab:app:ablation2}
\end{table}

\subsection{Two-stage experiment for coin tossing}
\label{s:app:two_stage_coin_tossing}

\begin{figure}[htpb]
\centering
\includegraphics[width=0.3\linewidth]{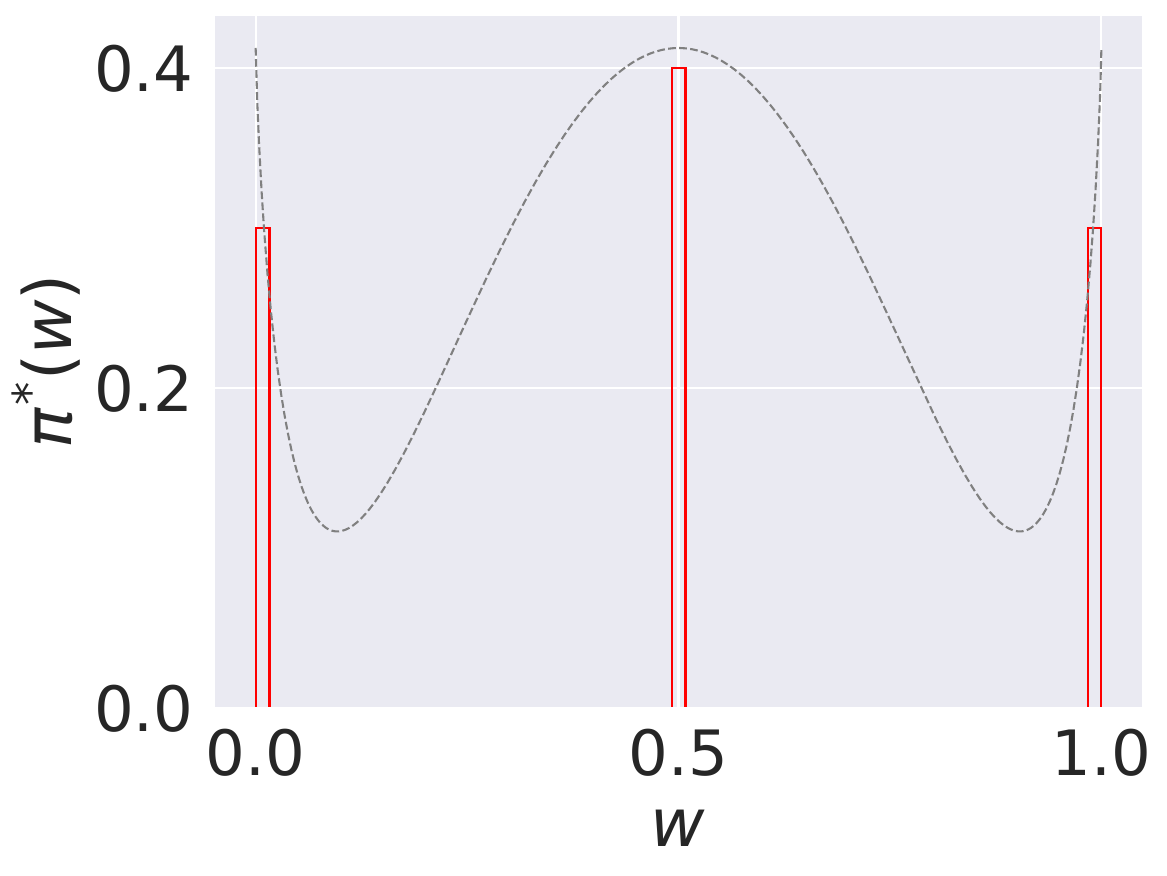}
\includegraphics[width=0.3\linewidth]{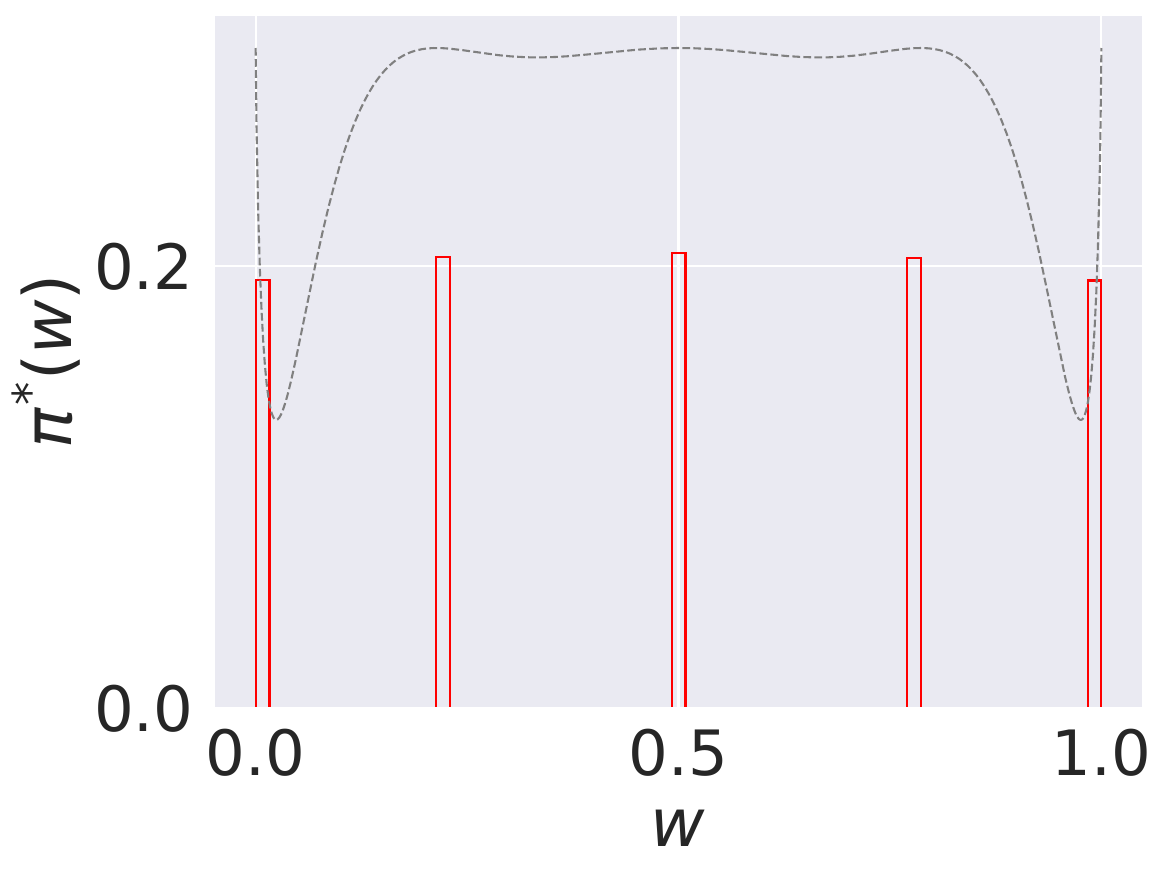}
\includegraphics[width=0.3\linewidth]{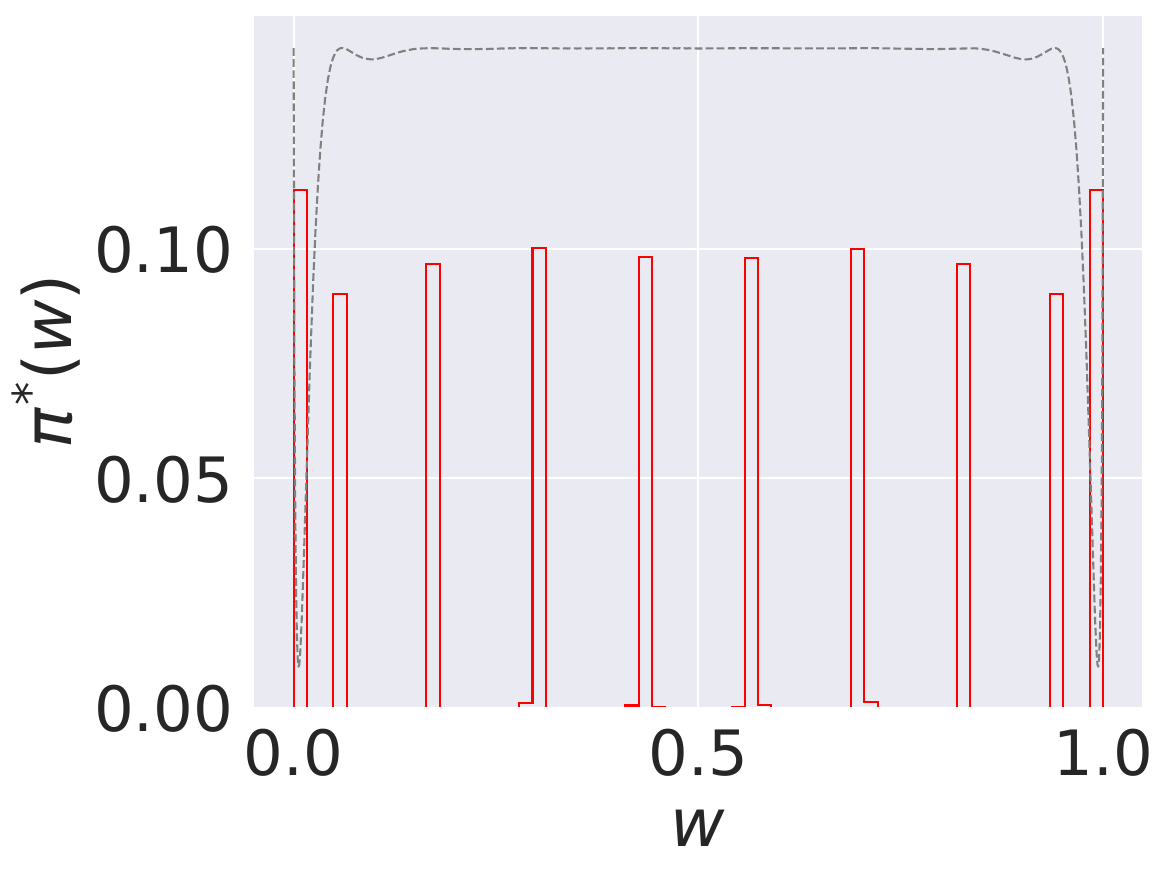}
\caption{\textbf{Reference prior for the two stage coin-tossing model (see~\cref{eq:ref_pinm_avg})} for $m=1$ and $n=1, 10, 40$ (from left to right) computed using the Blahut-Arimoto algorithm. Atoms are critical points of the gray line which is $\KL(p(\zmn), p(\zmn \mid w))- \KL(p(\zm), p(\zm \mid w))$. The prior is again discrete for finite order $n < \infty$. We see how this reference prior behaves for different values of $\a = m/n$, e.g., for $\a \to 0$ this prior $\pi^*$ is close to $\pin^*$ in~\cref{eq:ref_pin} but there are still some differences between them. This shows that the two-stage reference prior is not the same as the single-stage reference prior.}
\label{fig:app:coin_two_stage1}
\end{figure}

\begin{figure}[!htb]
\centering
\includegraphics[width=0.3\linewidth]{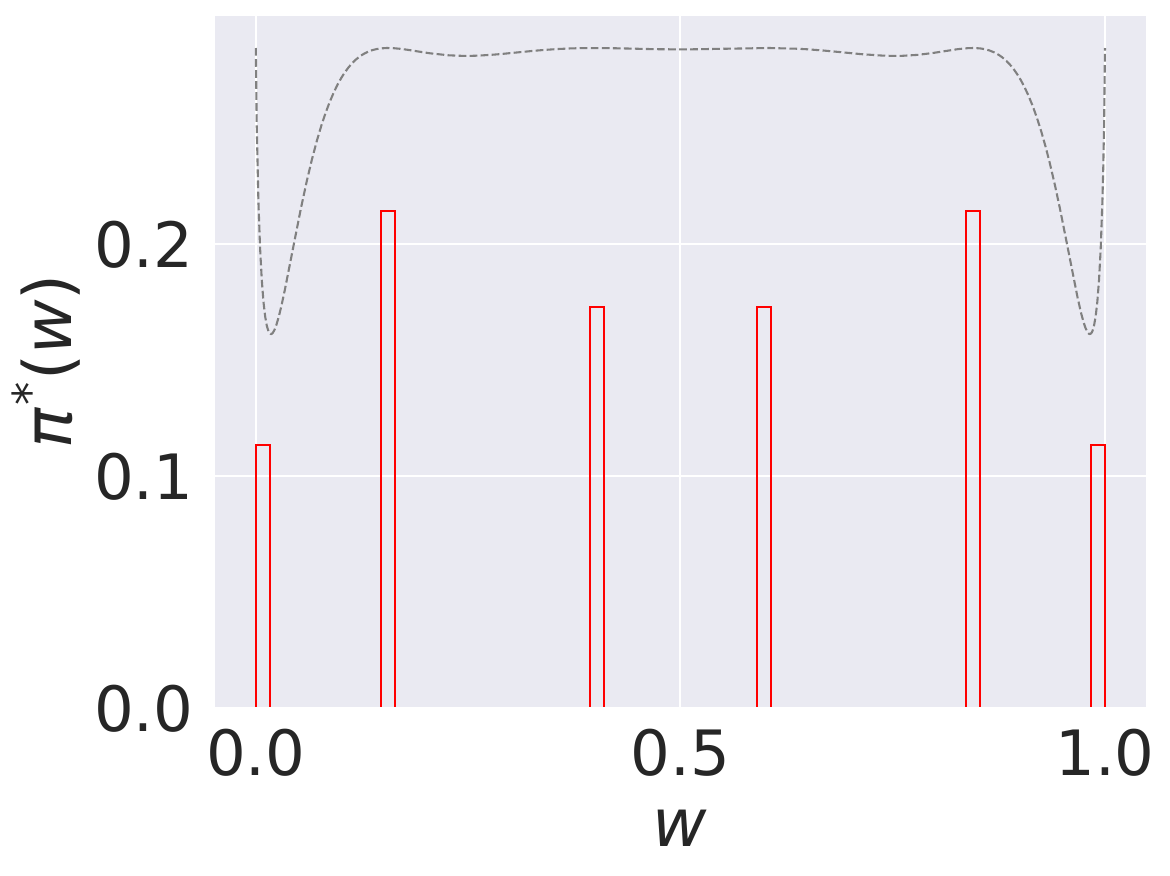}
\includegraphics[width=0.3\linewidth]{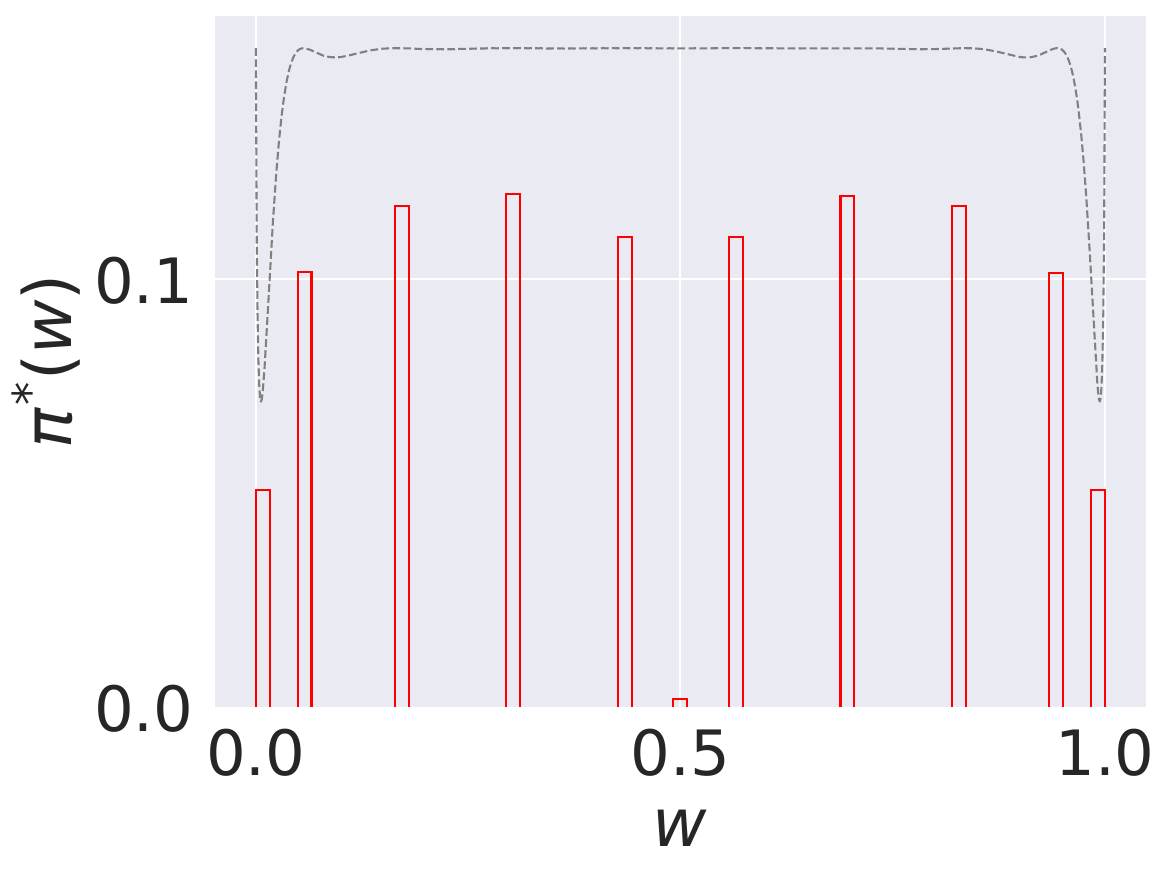}
\caption{\textbf{Reference prior for the two stage coin-tossing model} (see~\cref{eq:ref_pinm_avg}) for $n=1$ and $m=10, 30$ (from left to right) computed using the Blahut-Arimoto algorithm. Atoms are critical points of the gray line which is $\KL(p(\zmn), p(\zmn \mid w))- \KL(p(\zm), p(\zm \mid w))$.}
\label{fig:app:coin_two_stage2}
\end{figure}

In~\cref{s:two_stage}, we consider a situation when we obtain data in two stages, first $\zm$, and then $\zn$. We propose a prior $\pi^*$~\cref{eq:ref_pinm_avg} such that the posterior of the second stage makes the maximal use of the new $n$ samples. In this section, we visualize $\pi^*$ in the parameter space using a two-stage coin tossing experiment.
Consider the estimation of the bias of a coin $w \in [0,1]$ using two-stage $m+n$ trials. There are $m$ trials in first stage and $n$ trails in second stage. If $z$ denotes the number of heads in total, we have $p(z \mid w) = w^z (1-w)^{m+n-z} (m+n)!/(z! (m+n-z)!)$. We numerically find $\pi^*$ for different values of $m$ and $n$ using the BA algorithm (\cref{fig:app:coin_two_stage1} and \cref{fig:app:coin_two_stage2}).

